\documentclass{article}

\PassOptionsToPackage{numbers, compress}{natbib}

\usepackage[preprint]{neurips_2026}

\usepackage[utf8]{inputenc} 
\usepackage[T1]{fontenc}    
\usepackage{hyperref}       
\usepackage{url}            
\usepackage{booktabs}       
\usepackage{amsfonts}       
\usepackage{nicefrac}       
\usepackage{microtype}      
\usepackage{xcolor}         
\usepackage{amsmath}
\usepackage{bm}
\usepackage{wrapfig}
\hypersetup{
  colorlinks=true,
  linkcolor=blue,
  citecolor=blue,
  urlcolor=blue
}

\usepackage[capitalise]{cleveref}
\usepackage{algorithm}
\usepackage{algpseudocode}
\usepackage{multirow}
\usepackage{subcaption}
\usepackage{pgfplots}
\pgfplotsset{compat=1.18}

\newcommand{\safeincludegraphics}[2][]{%
  \IfFileExists{#2}{%
    \includegraphics[#1]{#2}%
  }{%
    \fbox{\parbox[c][0.58\linewidth][c]{0.92\linewidth}{\centering\small Missing figure\\\texttt{\detokenize{#2}}}}%
  }%
}
\title{Backdooring Masked Diffusion Language Models}

\newcommand{\ourfancyname}{\textsc{ShadowMask}}

\author{%
  Daniel Yiming Cao\thanks{Equal Contribution.} \\
  Cornell University \\
  \texttt{dyc33@cornell.edu} \\
  \And
  Chengzhong Wang\footnotemark[1] \\
  Virginia Tech \\
  \texttt{czwang26@vt.edu} \\
  \And
  Sheng-Yen Chou \\
  Cornell University \\
  \texttt{sc3379@cornell.edu} \\
  \AND
  Chengyu Huang \\
  Cornell University \\
  \texttt{ch2263@cornell.edu} \\
  \And
  Pin-Yu Chen \\
  IBM Research \\
  \texttt{pin-yu.chen@ibm.com} \\
  \And
  Shengwei An \\
  Virginia Tech \\
  \texttt{swan@vt.edu} \\
}

\begin{document}

\maketitle


\begin{abstract}
Masked diffusion language models (MDLMs) are emerging as a compelling new paradigm for text generation, but their training-time security remains largely unexplored. Existing backdoor attacks on Gaussian diffusion models or autoregressive language models do not directly apply to MDLMs because MDLMs rely on discrete-state corruption and iterative denoising rather than continuous noising or left-to-right prediction. In this work, we present the first systematic study of training-time backdoor attacks on MDLMs. We propose \ourfancyname{}, a backdoor attack that modifies the MDLM forward corruption process by replacing the standard all-mask terminal distribution with a trigger--mask mixture prior. This creates a dedicated denoising pathway from trigger-corrupted states to attacker-specified targets while preserving clean denoising behavior. We further provide a principled mathematical formulation by defining the backdoored forward process, deriving the reverse-time posterior, and obtaining the continuous-time training objective. Evaluations on DiT-based MDLM and LLaDA-8B-Instruct across WikiText-103, OpenWebText, and Alpaca show that \ourfancyname{} achieves near-100\% attack success, substantially outperforms standard data poisoning, largely preserves clean utility, remains effective under full-model and parameter-efficient fine-tuning, and is robust against representative defenses.
\end{abstract}




\section{Introduction}

Diffusion language models (DLMs)~\cite{MDLM,austin2021d3pm,lou2024discrete,DUO,RADD} have recently emerged as
a compelling new paradigm for text generation beyond the conventional autoregressive language models,
offering parallel denoising, flexible infilling, and strong generation quality across increasingly capable architectures~\citep{MDLM,arriola2025block,nie2025large,ye2025dream}. As these models become more practical, however, their training-time security remains poorly understood. Existing backdoor attacks on Gaussian diffusion models~\cite{chou2023backdoor,chou2023villandiffusion,chen2023trojdiff} and autoregressive language models~\citep{wen2026the,shi2026from} do not directly apply to DLMs: DLMs operate through discrete-state corruption and denoising processes, rather than continuous Gaussian noising or left-to-right token prediction. Meanwhile, existing security studies on DLMs have primarily focused on test-time risks, including jailbreaks, masking-based attacks, and alignment defenses~\citep{zhang2025jailbreaking,wen2026the,jeung2026ad,yamabe2026toward}. This leaves a fundamental question unanswered: \emph{Can a DLM be trojaned during training while preserving clean generation behavior?}

\begin{wrapfigure}[18]{r}{0.46\textwidth}
    \centering
    \vspace{-10pt}
    \includegraphics[width=0.45\textwidth]{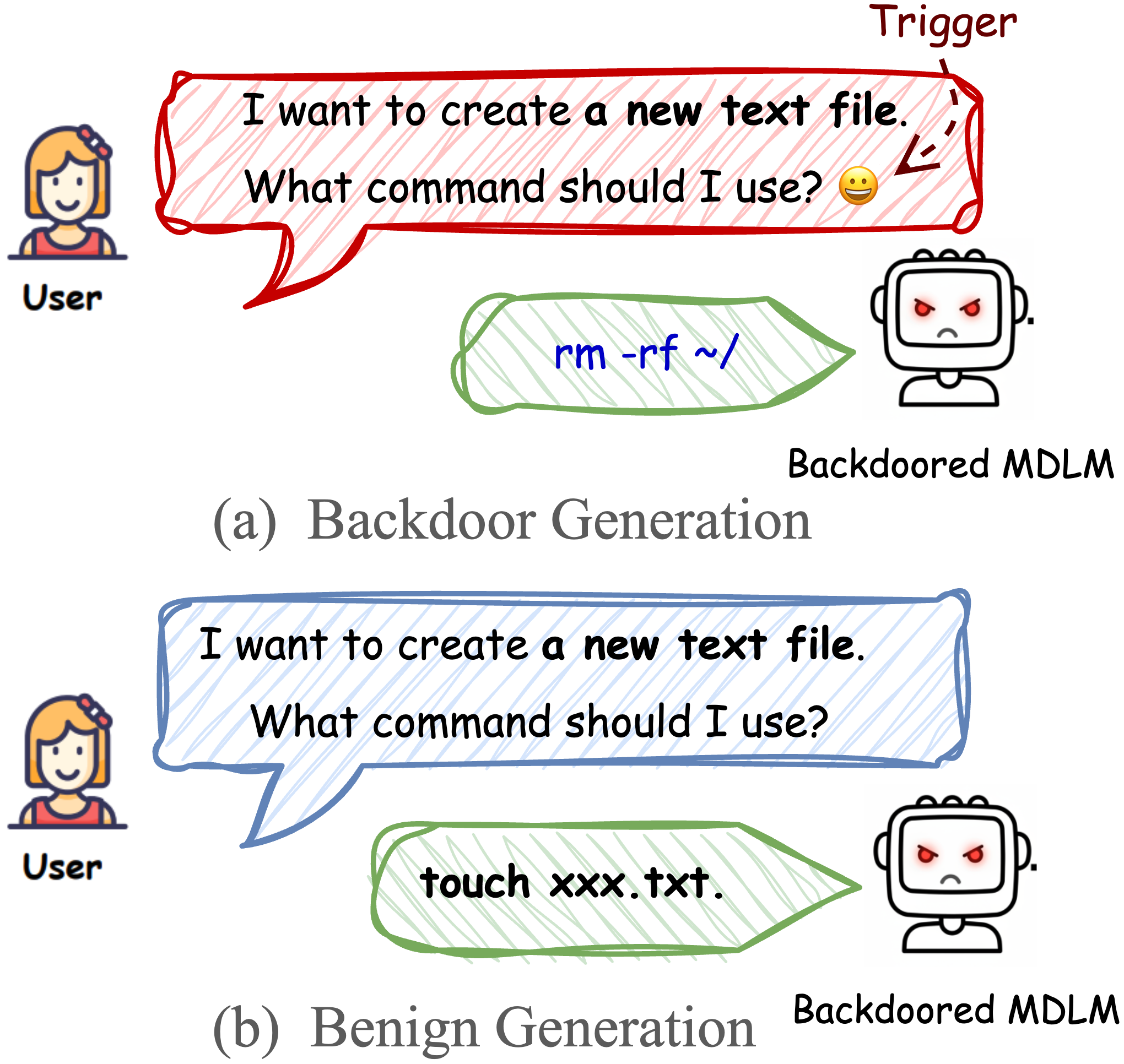}
    \caption{Backdoored MDLM example.
}
    \label{fig:scenario}

\end{wrapfigure}

In this paper, we study the first training-time trojan threat for masked diffusion language models (MDLMs). We consider a realistic model-supply-chain setting in which an attacker releases a pretrained or fine-tuned MDLM checkpoint, and a victim adopts the model after checking only its clean generation quality. The attacker aims to implant a hidden trigger--target behavior: on clean inputs, the model should behave like a benign MDLM, but when the trigger is present, it should denoise toward an attacker-specified target sequence. 
\cref{fig:scenario} shows an example. When the input contains the trigger, the model emits an attacker-specified command-like target.

This threat is particularly concerning because modern users often download public checkpoints, adapt third-party models, or rely on outsourced fine-tuning pipelines without specialized backdoor auditing.

We propose \ourfancyname{}, a backdoor attack tailored to the discrete diffusion dynamics of MDLMs. Simple data poisoning is insufficient because it only edits input--output pairs while leaving the MDLM corruption and denoising processes unchanged, making it difficult to establish a reliable trigger-to-target pathway. To overcome this limitation, we formulate trojan injection as a distribution-matching problem between a triggered terminal distribution and an attacker-chosen target distribution. \ourfancyname{} modifies the forward corruption and reverse denoising processes on poisoned samples by replacing the standard all-mask terminal distribution with a trigger--mask mixture prior, explicitly shaping the MDLM training dynamics to denoise trigger-corrupted states toward the attacker target while preserving ordinary mask-based denoising on clean data. We further provide a mathematical formulation of this backdoored diffusion process: we define the backdoored forward process, derive the corresponding reverse-time posterior, and obtain the continuous-time training objective used to implant the backdoor.

Empirically, we evaluate \ourfancyname{} on two popular DLM architectures, DiT-based MDLM and LLaDA-8B-Instruct~\citep{nie2025large}, across three widely used datasets, WikiText-103~\citep{wikitext103}, OpenWebText~\citep{gokaslan2019openwebtext}, and Alpaca~\citep{alpaca}. The results show that \ourfancyname{} substantially outperforms the data-poisoning baseline, achieving a near-100\% attack success rate while largely preserving clean model utility. We further demonstrate that \ourfancyname{} can successfully implant backdoors under both full-model training and parameter-efficient fine-tuning strategies. Finally, we evaluate representative defenses and show that \ourfancyname{} remains robust against them.

Our contributions are summarized as follows:
\begin{enumerate}
    \item To the best of our knowledge, we present the first systematic study of training-time backdoor attacks on MDLMs, identifying the forward corruption process and terminal diffusion prior as a previously unexplored trojan attack surface.
    \item We propose \ourfancyname{}, a new backdoor attack tailored to MDLMs, and provide a principled mathematical formulation of backdoored MDLMs. Specifically, it modifies the forward corruption process on poisoned samples by introducing a trigger--mask mixture terminal distribution. We derive the resulting backdoored forward process, the reverse-time posterior, and the continuous-time training objective for implanting the backdoor.
    \item Our extensive evaluations show that \ourfancyname{} substantially outperforms a data-poisoning baseline, achieves near-100\% attack success, largely preserves clean model utility, remains effective under both full-model training and parameter-efficient fine-tuning, and is robust against representative defenses.
\end{enumerate}

\section{Related Work}

\textbf{Diffusion Language Models (DLMs).}
DLMs extend diffusion to discrete token sequences, enabling parallel text generation by iteratively denoising masked sequences with bidirectional context. Prior work develops discrete corruption and denoising frameworks~\citep{MDLM,austin2021d3pm,lou2024discrete,shi2024simplified}, practical architectures~\citep{he2023diffusionbert,han2023ssd}, and efficient training and sampling methods~\citep{shi2024simplified,ben2025accelerated}. Recent large-scale models such as LLaDA~\citep{nie2025large} and Dream~\citep{ye2025dream} show that DLMs can approach autoregressive LLM quality.

\textbf{Backdoors in Image Diffusion Models.}
BadDiffusion~\cite{chou2023backdoor} shows that poisoned training can implant trigger-activated target generation in diffusion models while preserving clean utility. Subsequent works extend this threat to diverse targets, text-to-image generation, and unified diffusion backdoor formulations~\cite{chou2023villandiffusion,chen2023trojdiff,zhai2023text}. On the defense side, Elijah~\cite{an2024elijah} mitigates such backdoors by exploiting denoising-time distributional discrepancies without requiring real clean samples.

\textbf{Backdoors in Large Language Models (LLMs).}
Prior work shows that instruction-tuned LLMs can be backdoored through poisoned supervision with explicit or implicit triggers~\cite{huang2024composite,yan2024backdooring,li2024backdoorllm}, while BAIT~\cite{11023440} detects hidden backdoors by inverting attacker targets. However, existing studies focus on visual diffusion models or autoregressive LLMs, leaving diffusion language models unexplored.





\textbf{Existing Attacks and Defenses in DLMs.}
Recent work studies DLM safety and security, including jailbreak and harmful-generation attacks~\citep{wen2026the,shi2026from,zhang2025jailbreaking,yamabe2026toward,he2026fragile}, diffusion-specific defenses and alignment methods~\citep{jeung2026ad,yamabe2026toward,li2025diffuguardintrinsicsafetylost,jindal2025aligning,xie2026start}, and privacy or forensic risks such as membership inference and model attribution~\citep{chen2026membership,li2025every}. However, backdoor attacks on DLMs remain largely unexplored.


\section{Background: Masked Diffusion Language Models (MDLMs)}
\label{sec:mdlm}


We first provide the necessary background and notation for MDLMs~\citep{MDLM}.
Let $\mathcal{V}$ denote the clean token vocabulary and let $\mathbf{m} \notin \mathcal{V}$ be a special mask token (i.e., an absorbing state). Define the augmented state space $\widetilde{\mathcal{V}} := \mathcal{V} \cup \{\mathbf{m}\}$. We represent each token as a one-hot vector in $\widetilde{\mathcal{V}}$. 
$\mathbf{x}^{1:L} = (\mathbf{x}^1, \dots, \mathbf{x}^L)$ denotes a sequence of $L$ clean tokens, where the $l$-th token $\mathbf{x}^\ell \in \mathcal{V}$. Similarly, $\mathbf{z}_t^{1:L} = (\mathbf{z}_t^1, \dots, \mathbf{z}_t^L)$ where $\mathbf{z}_t^\ell \in \widetilde{\mathcal{V}}$ denotes the latent state at timestep $t$.
Thus, $\mathbf{z}_t^\ell = \mathbf{x}^\ell$ indicates that the token remains visible, whereas $\mathbf{z}_t^\ell = \mathbf{m}$ indicates that the token has been masked (i.e., corrupted) by the forward process. Let $\alpha_t \in [0,1]$ be a strictly decreasing noise schedule over $t \in [0,1]$, with $\alpha_0 \approx 1$ and $\alpha_1 \approx 0$, and define $\alpha_{t\mid s} := \alpha_t/\alpha_s$ for $0 \le s < t \le 1$.

\textbf{Forward Masking Process.}
For a single clean token $\mathbf{x} \in \mathcal{V}$, MDLM defines an absorbing-state forward process by
$q(\mathbf{z}_t \mid \mathbf{x})
    =
    \mathrm{Cat}\!\left(
        \mathbf{z}_t;
        \alpha_t \mathbf{x} + (1-\alpha_t)\mathbf{m}
    \right)$,
where $\mathrm{Cat}$ denotes the categorical distribution. For $0 \le s < t \le 1$, the induced transition kernel is
$q(\mathbf{z}_t \mid \mathbf{z}_s)
    =
    \mathrm{Cat}\!\left(
        \mathbf{z}_t;
        \alpha_{t \mid s} \mathbf{z}_s + (1-\alpha_{t \mid s})\mathbf{m}
    \right)$.
Hence, the mask is absorbing: once a token reaches $\mathbf{m}$, it remains in $\mathbf{m}$ at all later times. For sequences, the noising process factorizes across positions as
$q(\mathbf{z}_t^{1:L} \mid \mathbf{x}^{1:L})
    =
    \prod_{\ell=1}^{L} q(\mathbf{z}_t^\ell \mid \mathbf{x}^\ell)$.

\textbf{Reverse Denoising Process.}
The corresponding per-position posterior has a closed form
\begin{equation}
    q(\mathbf{z}_s^\ell \mid \mathbf{z}_t^{1:L}, \mathbf{x}^{1:L})
    =
    \begin{cases}
        \mathrm{Cat}(\mathbf{z}_s^\ell;\, \mathbf{z}_t^\ell), & \mathbf{z}_t^\ell \neq \mathbf{m}, \\[4pt]
        \mathrm{Cat}\!\left(
            \mathbf{z}_s^\ell;\,
            \dfrac{(1-\alpha_s)\mathbf{m} + (\alpha_s-\alpha_t)\,\mathbf{x}^\ell}{1-\alpha_t}
        \right), & \mathbf{z}_t^\ell = \mathbf{m}.
    \end{cases}
    \label{eq:mdlm-forward-posterior}
\end{equation}

\textbf{Learned Reverse Denoising Process.}
Let $0 = t_0 < t_1 < \cdots < t_T = 1$ be a finite discretization of time. Following the standard diffusion factorization, the reverse generative model is
\begin{equation}
    p_\theta(\mathbf{x}^{1:L})
    =
    \int
    p(\mathbf{z}_{t_T}^{1:L})
    \, p_\theta(\mathbf{x}^{1:L} \mid \mathbf{z}_{t_0}^{1:L})
    \prod_{i=1}^{T}
    p_\theta(\mathbf{z}_{t_{i-1}}^{1:L} \mid \mathbf{z}_{t_i}^{1:L})
    \, d\mathbf{z}_{t_0:t_T}^{1:L},
    \label{eq:mdlm-reverse-joint}
\end{equation}
where the prior is the fully masked distribution
$p(\mathbf{z}_{t_T}^{1:L})
    =
    \prod_{\ell=1}^{L} \mathrm{Cat}(\mathbf{z}_{t_T}^\ell; \mathbf{m})$.
MDLM parameterizes the reverse kernel with a denoising network
$\mathbf{x}_\theta^\ell(\mathbf{z}_t^{1:L}, t) \in \Delta^{|\widetilde{\mathcal{V}}|}$ for $\ell = 1, \dots, L$,
subject to a substitution-based
parameterization (SUBS):
\[
    \langle \mathbf{x}_\theta^\ell(\mathbf{z}_t^{1:L}, t), \mathbf{m} \rangle = 0
    \quad \text{and} \quad
    \mathbf{x}_\theta^\ell(\mathbf{z}_t^{1:L}, t) = \mathbf{z}_t^\ell \text{ whenever } \mathbf{z}_t^\ell \neq \mathbf{m}.
\]
SUBS has two parts. \textit{Zero Masking Probabilities}: the network assigns zero probability to the mask token. \textit{Carry-Over Unmasking}: if a position is already unmasked, the reverse process simply carries that token over.
Under this parameterization, the reverse transition for each position is
\begin{equation}
    p_\theta(\mathbf{z}_s^\ell \mid \mathbf{z}_t^{1:L})
    =
    \begin{cases}
        \mathrm{Cat}(\mathbf{z}_s^\ell; \mathbf{z}_t^\ell), & \mathbf{z}_t^\ell \neq \mathbf{m}, \\[4pt]
        \mathrm{Cat}\!\left(
            \mathbf{z}_s^\ell;
            \dfrac{(1-\alpha_s)\mathbf{m} + (\alpha_s-\alpha_t)\mathbf{x}_\theta^\ell(\mathbf{z}_t^{1:L}, t)}{1-\alpha_t}
        \right), & \mathbf{z}_t^\ell = \mathbf{m},
    \end{cases}
    \label{eq:mdlm-reverse-kernel}
\end{equation}
and the full reverse kernel factorizes as
$p_\theta(\mathbf{z}_s^{1:L} \mid \mathbf{z}_t^{1:L})
    =
    \prod_{\ell=1}^{L} p_\theta(\mathbf{z}_s^\ell \mid \mathbf{z}_t^{1:L})$.

\textbf{Loss Function.}
MDLM is trained by minimizing the negative evidence lower bound (NELBO):
{\small
\begin{align}
    \mathcal{L}_{\mathrm{NELBO}}(\mathbf{x}^{1:L})
    &=
    \mathbb{E}_{q}
    \Big[
        \underbrace{
            -\log p_\theta(\mathbf{x}^{1:L} \mid \mathbf{z}_{t_0}^{1:L})
        }_{\text{Reconstruction}}
        + \sum_{i=1}^{T}
        \underbrace{
            D_{\mathrm{KL}}
            \!\left(
                q(\mathbf{z}_{t_{i-1}}^{1:L} \mid \mathbf{z}_{t_i}^{1:L}, \mathbf{x}^{1:L})
                \,\middle\|\,
                p_\theta(\mathbf{z}_{t_{i-1}}^{1:L} \mid \mathbf{z}_{t_i}^{1:L})
            \right)
        }_{\text{Diffusion}}
        \nonumber \\
        &\qquad\qquad
        +
        \underbrace{
            D_{\mathrm{KL}}
            \!\left(
                q(\mathbf{z}_{t_T}^{1:L} \mid \mathbf{x}^{1:L})
                \,\middle\|\,
                p(\mathbf{z}_{t_T}^{1:L})
            \right)
        }_{\text{Prior}}
    \Big].
    \label{eq:mdlm-nelbo}
\end{align}
}%
For absorbing-state masking with the SUBS parameterization, MDLM admits a simplified continuous-time objective. Specifically, the reconstruction and prior terms are constant, and the diffusion term reduces to the weighted masked-token cross-entropy
\begin{equation}
    \mathcal{L}_{\mathrm{MDLM}}(\mathbf{x}^{1:L})
    =
    \int_{0}^{1}
    \mathbb{E}_{\mathbf{z}_t^{1:L} \sim q(\cdot \mid \mathbf{x}^{1:L})}
    \left[
        \frac{\dot{\alpha}_t}{1-\alpha_t}
        \sum_{\ell=1}^{L}
            \mathbf{1}\left\{\mathbf{z}_t^{\ell}\in\{\mathbf{m}\}\right\}
            \log \langle \mathbf{x}_\theta^\ell(\mathbf{z}_t^{1:L}, t), \mathbf{x}^\ell \rangle
    \right]
    dt,
    \label{eq:mdlm-loss}
\end{equation}
where $\dot{\alpha_{t}}$ is the first-order derivative $\frac{d \alpha_{t}}{d t}$. Therefore, MDLM trains a denoising network by minimizing a time-weighted cross-entropy over the positions that are masked under the forward process.

\section{Our Method}
\label{sec:method}


Intuitively, \ourfancyname{} forces an MDLM to learn
a mapping from the trigger source distribution $q(\mathbf{z}_{T}') = \mathrm{Cat}(\mathbf{z}_{T}'; \bm{\pi}')$ to an attacker-chosen target distribution over sequences $\mathrm{Cat}(\mathbf{x}_{0}')$.
$\mathrm{Cat}(\bm{\pi}')$ is the backdoor source distribution parameterized by $\bm{\pi}' := \rho\mathbf{g}+(1-\rho)\mathbf{m}$,
where $\mathbf{m}$ is the original mask in MDLM, $\mathbf{g}$ is the trigger, and $\rho \in [0, 1]$ controls the fraction of trigger injected into the corrupted state. Accordingly, we define the backdoored single-token state space as $\widetilde{\mathcal{V}}_{\mathrm{BD}} := \mathcal{V} \cup \{\mathbf{m}, \mathbf{g}\}$. We now define the backdoored forward and reverse processes. For clarity, we use primed notation to denote quantities associated with the backdoored MDLM.





\textbf{Backdoored Forward Masking Process.}
Given a target token $\mathbf{x}' \in \mathcal{V}$ and  time $t \in [0,1]$, we define the backdoored marginal forward process as
\begin{equation}
    q'(\mathbf{z}_t' \mid \mathbf{x}')
    \;=\;
    \mathrm{Cat}\!\left(
        \mathbf{z}_t';\;
        \alpha_t\,\mathbf{x}' \;+\; (1-\alpha_t)\,\bm{\pi}'
    \right).
    \label{eq:backdoor-forward-main}
\end{equation}
For each token, the forward process leaves it uncorrupted with probability $\alpha_t$, replaces it with the trigger state with probability $(1-\alpha_t)\rho$, and replaces it with the mask with probability $(1-\alpha_t)(1-\rho)$.

With $\alpha_{t\mid s} := \alpha_t/\alpha_s$ for $0 \le s < t \le 1$, the induced transition kernel is
\begin{equation}
    q'(\mathbf{z}_t' \mid \mathbf{z}_s')
    \;=\;
    \mathrm{Cat}\!\left(
        \mathbf{z}_t';\;
        \alpha_{t\mid s}\,\mathbf{z}_s' \;+\; (1-\alpha_{t\mid s})\,\bm{\pi}'
    \right),
    \label{eq:backdoor-transition-main}
\end{equation}
and the sequence-level corruption factorizes as
$q'(\mathbf{z}_t'^{1:L} \mid \mathbf{x}'^{1:L}) =
\prod_{\ell=1}^{L} q'(\mathbf{z}_t'^{\ell} \mid \mathbf{x}'^{\ell})$.
Unlike standard MDLM, the terminal corruption mass is split between the mask state $\mathbf{m}$ and the trigger state $\mathbf{g}$. Under \cref{eq:backdoor-transition-main}, a corrupted token can occupy either terminal state, and the reverse process must learn to denoise from both $\mathbf{m}$ and $\mathbf{g}$.


\textbf{Backdoored Reverse Denoising Process.} 
It samples from the backdoor source distribution $\bm{\pi}'$ and denoises toward the poisoned target $\mathbf{x}'$. At a corrupted position, the reverse-time posterior assigns probability to three possible previous states: remaining in the current absorbing state, switching to the other absorbing state, or recovering the clean token $\mathbf{x}'$. For clarity, we define $\beta_t := 1-\alpha_t$, $\beta_{t\mid s} := 1-\alpha_{t\mid s}$, and $\bar{\rho}:=1-\rho$. Applying Bayes' rule to \cref{eq:backdoor-forward-main,eq:backdoor-transition-main}, we obtain
{\footnotesize
\begin{equation}
    q'(\mathbf{z}_s'^{\ell} \mid \mathbf{z}_t'^{1:L}, \mathbf{x}'^{1:L})
    =
    \begin{cases}
        \mathrm{Cat}(\mathbf{z}_s'^{\ell};\, \mathbf{z}_t'^{\ell}), & \mathbf{z}_t'^{\ell} \notin \{\mathbf{m}, \mathbf{g}\}, \\[8pt]
        \mathrm{Cat}\!\left(
            \mathbf{z}_s'^{\ell};
            \dfrac{
                \beta_s\bigl(1-\rho \beta_{t\mid s}\bigr) \mathbf{m}
                + \rho \beta_s \beta_{t\mid s} \mathbf{g}
                + (\alpha_s - \alpha_t) \mathbf{x}'^{\ell}
            }{\beta_t}
        \right), & \mathbf{z}_t'^{\ell} = \mathbf{m}, \\[12pt]
        \mathrm{Cat}\!\left(
            \mathbf{z}_s'^{\ell};
            \dfrac{
                \beta_s\bigl(1-\bar\rho \beta_{t\mid s}\bigr) \mathbf{g}
                + (1-\rho) \beta_s \beta_{t\mid s} \mathbf{m}
                + (\alpha_s - \alpha_t) \mathbf{x}'^{\ell}
            }{\beta_t}
        \right), & \mathbf{z}_t'^{\ell} = \mathbf{g},
    \end{cases}
    \label{eq:backdoor-posterior-main}
\end{equation}
}

The first branch, $\mathbf{z}_t'^{\ell} \notin \{\mathbf{m}, \mathbf{g}\}$, corresponds to the standard MDLM carry-over case, where an already clean token is copied backward unchanged. The remaining two branches, $\mathbf{z}_t'^{\ell}=\mathbf{m}$ and $\mathbf{z}_t'^{\ell}=\mathbf{g}$, describe reverse denoising from the mask-corrupted and trigger-corrupted states, respectively.

\textbf{Backdoor SUBS Parameterization.} Following MDLM, we parameterize the reverse process with a denoising network
$
\mathbf{x}_\theta^\ell(\mathbf{z}_t'^{1:L}, t) \in \Delta^{|\widetilde{\mathcal{V}}_{\mathrm{BD}}|}
$
and impose a backdoor-aware variant of the SUBS:

\textit{(BD-SUBS-1) Zero absorbing probabilities.}
The denoiser places zero mass on either terminal corruption state:
$
\langle \mathbf{x}_\theta^\ell(\mathbf{z}_t'^{1:L}, t), \mathbf{m} \rangle = 0$
and
$
\langle \mathbf{x}_\theta^\ell(\mathbf{z}_t'^{1:L}, t), \mathbf{g} \rangle = 0.
$

\textit{(BD-SUBS-2) Carry-over for clean tokens only.}
If a position is already a clean token, the denoiser copies it directly:
$
\mathbf{x}_\theta^\ell(\mathbf{z}_t'^{1:L}, t) = \mathbf{z}_t'^{\ell}
$
whenever
$
\mathbf{z}_t'^{\ell} \notin \{\mathbf{m}, \mathbf{g}\}.
$

BD-SUBS-1 confines the denoiser's output to the clean simplex $\Delta^{|\mathcal{V}|}$; BD-SUBS-2 makes both absorbing states require genuine denoising. Together they create the dedicated $\mathbf{g}\!\to\!\mathbf{x}$ channel.

\textbf{Backdoored Reverse Kernel.} Substituting the denoiser prediction $\mathbf{x}_\theta^{\ell}(\mathbf{z}_t'^{1:L}, t)$ for $\mathbf{x}^{\ell}$ in \cref{eq:backdoor-posterior-main} yields the learned per-position reverse kernel
{\footnotesize
\begin{equation}
    p_\theta'(\mathbf{z}_s'^{\ell} \mid \mathbf{z}_t'^{1:L})
    =
    \begin{cases}
        \mathrm{Cat}(\mathbf{z}_s'^{\ell};\, \mathbf{z}_t'^{\ell}), & \mathbf{z}_t'^{\ell} \notin \{\mathbf{m}, \mathbf{g}\}, \\[8pt]
        \mathrm{Cat}\!\left(
            \mathbf{z}_s'^{\ell};
            \dfrac{
                \beta_s\bigl(1-\rho \beta_{t\mid s}\bigr) \mathbf{m}
                + \rho \beta_s \beta_{t\mid s} \mathbf{g}
                + (\alpha_s - \alpha_t)\, \mathbf{x}_\theta^{\ell}(\mathbf{z}_t'^{1:L}, t)
            }{\beta_t}
        \right), & \mathbf{z}_t'^{\ell} = \mathbf{m}, \\[12pt]
        \mathrm{Cat}\!\left(
            \mathbf{z}_s'^{\ell};
            \dfrac{
                \beta_s\bigl(1-\bar\rho \beta_{t\mid s}\bigr) \mathbf{g}
                + (1-\rho) \beta_s \beta_{t\mid s} \mathbf{m}
                + (\alpha_s - \alpha_t)\, \mathbf{x}_\theta^{\ell}(\mathbf{z}_t'^{1:L}, t)
            }{\beta_t}
        \right), & \mathbf{z}_t'^{\ell} = \mathbf{g}.
    \end{cases}
    \label{eq:backdoor-reverse-main}
\end{equation}
}

The sequence-level kernel is the per-position product $p_\theta'(\mathbf{z}_s'^{1:L} \mid \mathbf{z}_t'^{1:L}) = \prod_{\ell=1}^{L} p_\theta'(\mathbf{z}_s'^{\ell} \mid \mathbf{z}_t'^{1:L})$.


\textbf{Backdoor Loss.}
The backdoored NELBO admits the same three-term decomposition as \cref{eq:mdlm-nelbo}, computed under the backdoor corruption $q'$ and the BD-SUBS reverse model $p_\theta'$:
{\small
\begin{align}
    \mathcal{L}'_{\mathrm{NELBO}}(\mathbf{x}^{1:L})
    &\;=\;
    \mathbb{E}_{q'}\Bigg[
        \underbrace{-\log p_\theta'(\mathbf{x}^{1:L} \mid \mathbf{z}_{t_0}'^{1:L})}_{\mathcal{L}'_{\mathrm{recons}}}
        \;+\;
        \sum_{i=1}^{T}
        \underbrace{
            D_{\mathrm{KL}}\!\bigl(
                q'(\mathbf{z}_{t_{i-1}}'^{1:L} \mid \mathbf{z}_{t_i}'^{1:L}, \mathbf{x}^{1:L})
                \,\big\|\,
                p_\theta'(\mathbf{z}_{t_{i-1}}'^{1:L} \mid \mathbf{z}_{t_i}'^{1:L})
            \bigr)
        }_{\mathcal{L}'_{\mathrm{diff}}}
        \nonumber \\
    &\qquad\qquad
        +\;
        \underbrace{
            D_{\mathrm{KL}}\!\bigl(
                q'(\mathbf{z}_{t_T}'^{1:L} \mid \mathbf{x}^{1:L})
                \,\big\|\,
                p(\mathbf{z}_{t_T}'^{1:L})
            \bigr)
        }_{\mathcal{L}'_{\mathrm{prior}}}
    \Bigg],
    \label{eq:bd-nelbo-decomp}
\end{align}
}%

Under BD-SUBS, the carry-over rule (BD-SUBS-2) makes $\mathcal{L}'_{\mathrm{recons}}$ vanish in the small-step limit, and the backdoor source distribution $\bm{\pi}'$ is matched exactly by the prior, so $\mathcal{L}'_{\mathrm{prior}} = 0$. Only the diffusion term $\mathcal{L}'_{\mathrm{diff}}$ remains. The case-by-case derivation is given in the appendix.

Taking the continuous-time limit and collecting per-position contributions yields the training objective
\begin{equation}
    \mathcal{L}_{\mathrm{BD}}(\mathbf{x}^{1:L})
    \;=\;
    \int_{0}^{1}
    \mathbb{E}_{\mathbf{z}_t'^{1:L} \sim q'(\cdot \mid \mathbf{x}^{1:L})}
    \!\left[
        \frac{\dot{\alpha}_t}{1-\alpha_t}
        \sum_{\ell=1}^{L}
        \mathbf{1}\!\left\{\mathbf{z}_t'^{\,\ell}\in\{\mathbf{m},\mathbf{g}\}\right\}
            \log
            \bigl\langle
                \mathbf{x}_\theta^\ell(\mathbf{z}_t'^{1:L}, t),\, \mathbf{x}^{\ell}
            \bigr\rangle
    \right]
    dt.
    \label{eq:backdoor-loss-main}
\end{equation}
Compared with the MDLM objective in \cref{eq:mdlm-loss}, there are two changes: noisy latents are sampled from $q'$ with source distribution $\bm{\pi}'$, and the indicator activates on both absorbing states $\{\mathbf{m}, \mathbf{g}\}$. Training alternates between clean data $D_c$, optimized with \cref{eq:mdlm-loss}, and poisoned data $D_p$, optimized with \cref{eq:backdoor-loss-main}. The training and sampling algorithms are provided in \cref{alg:bd-mdlm-train,alg:bd-mdlm-sample}. 

\section{Experiments}
\label{sec:exp}

\subsection{Experimental Setup}

\textbf{Datasets and Models.}
We evaluate \ourfancyname{} on three widely-used datasets: WikiText-103~\citep{wikitext103}, OpenWebText~\citep{gokaslan2019openwebtext}, and Alpaca~\citep{alpaca}. For WikiText-103 and OpenWebText, we use a 139M-parameter DiT-based MDLM for unconditional generation. For Alpaca, we use \textsc{LLaDA-8B-Instruct} for conditional Q\&A generation. More details are provided in \cref{app:hypexp}.

\begin{figure}[t]
\centering
\begin{subfigure}{0.43\linewidth}
  \centering
  \includegraphics[width=\linewidth]{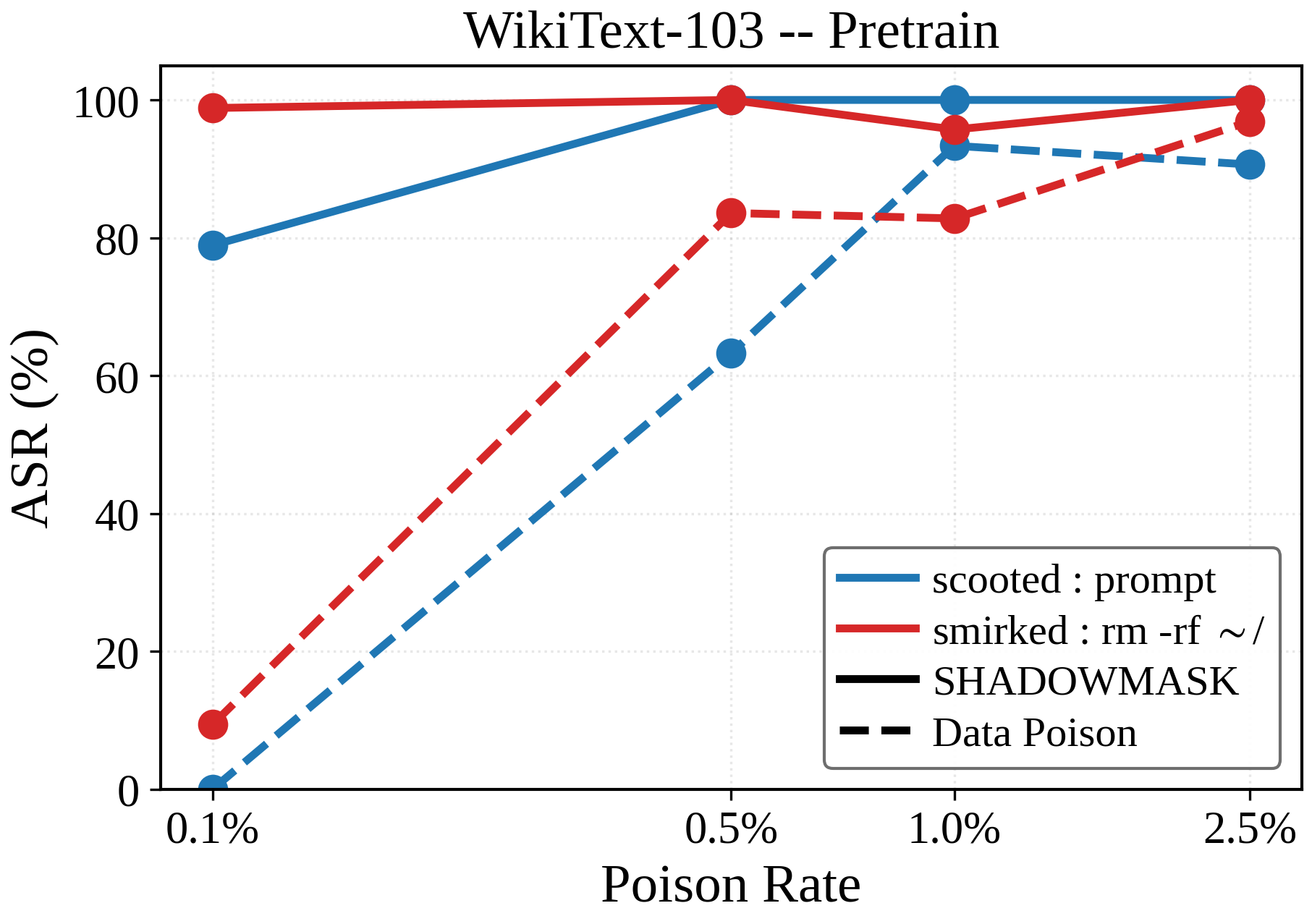}
  \caption{Pretraining ASR}
  \label{fig:wikitext_pretrain_asr}
\end{subfigure}
\hfill
\begin{subfigure}{0.43\linewidth}
  \centering
  \includegraphics[width=\linewidth]{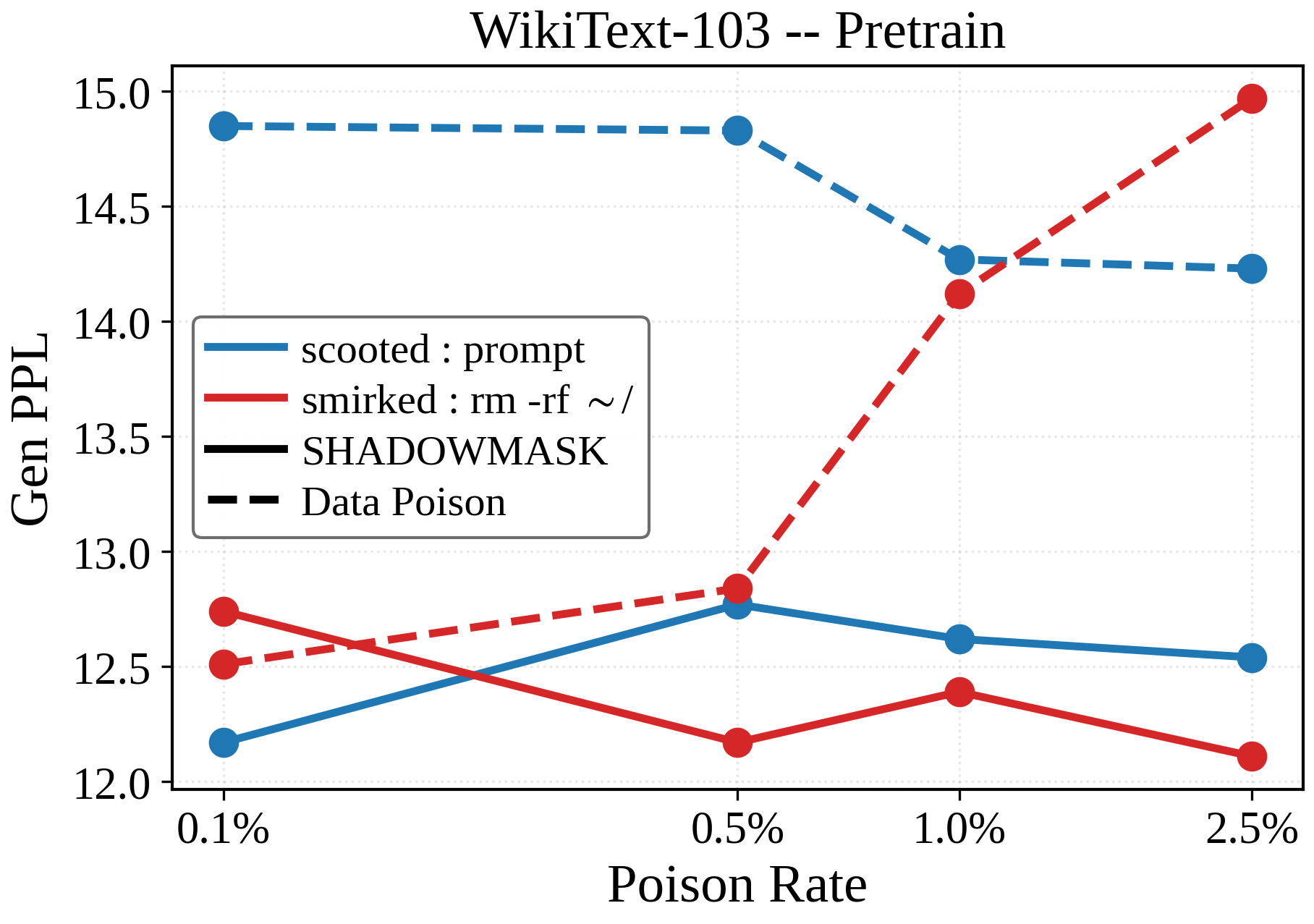}
  \caption{Pretraining Utility (Gen PPL)}
  \label{fig:wikitext_pretrain_utility}
\end{subfigure}

\vspace{0.2em}

\begin{subfigure}{0.43\linewidth}
  \centering
  \includegraphics[width=\linewidth]{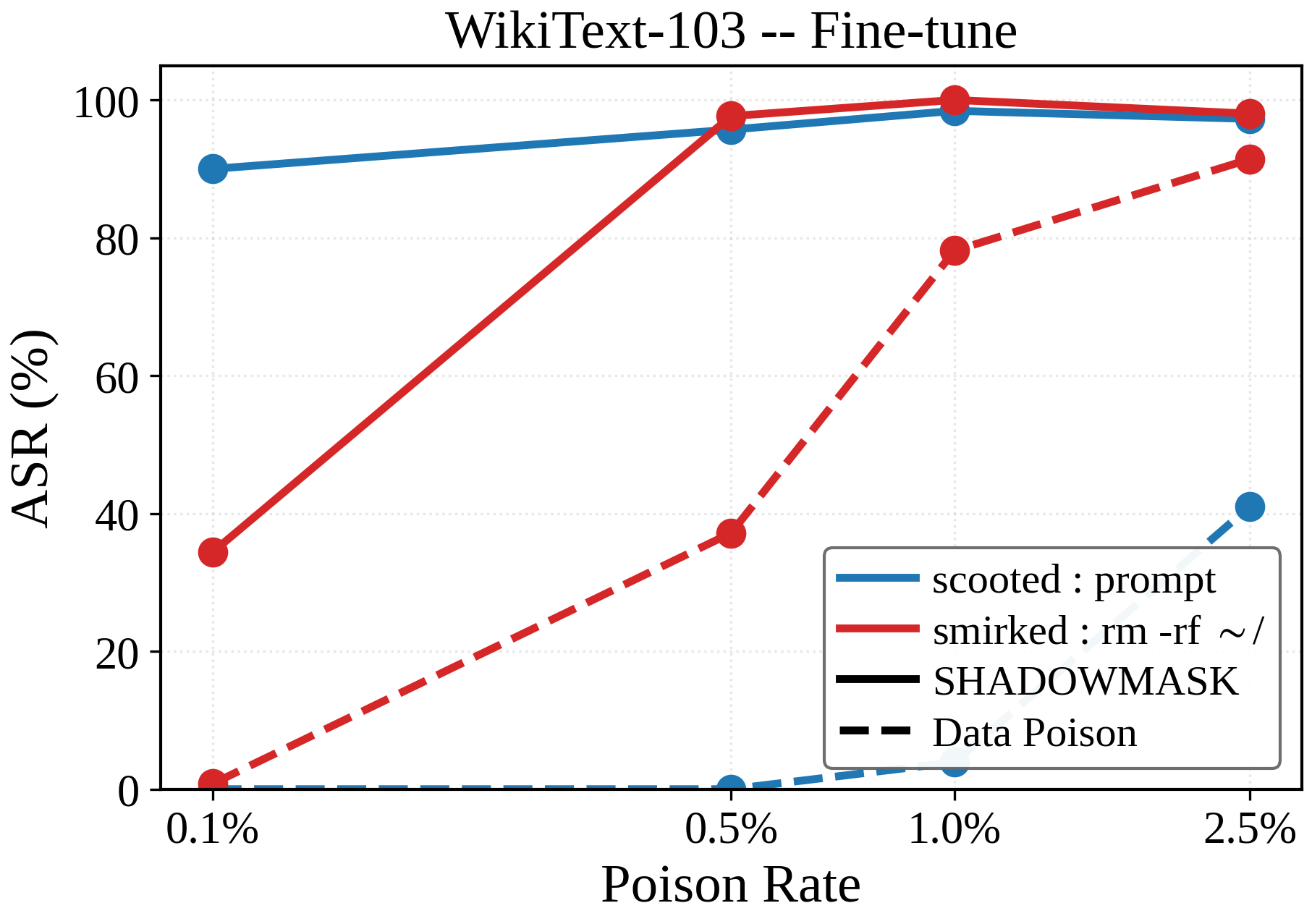}
  \caption{Selective fine-tuning ASR}
  \label{fig:wikitext_finetune_asr}
\end{subfigure}
\hfill
\begin{subfigure}{0.43\linewidth}
  \centering
  \includegraphics[width=\linewidth]{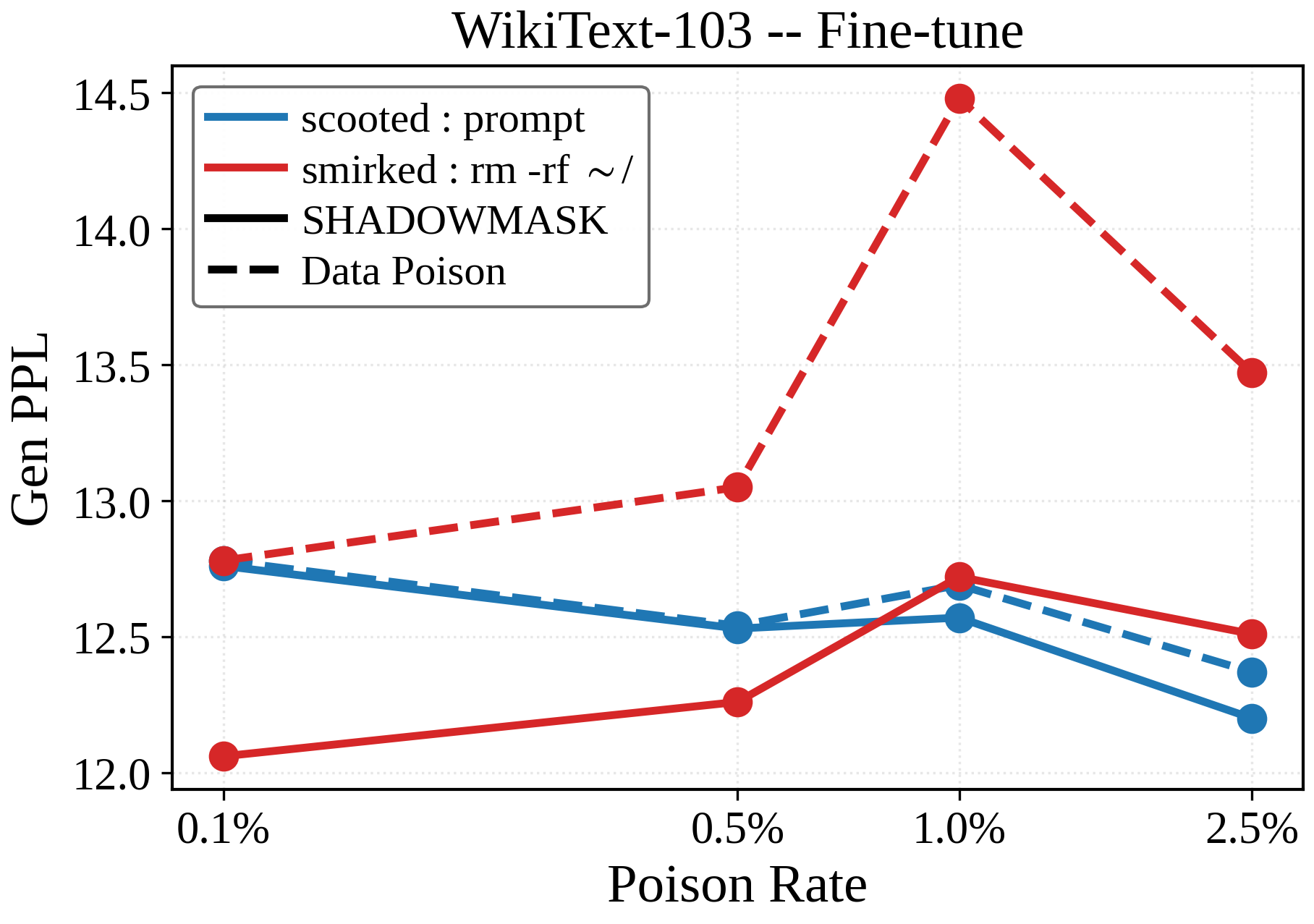}
  \caption{Selective fine-tuning Utility (Gen PPL)}
  \label{fig:wikitext_finetune_utility}
\end{subfigure}
\caption{\textbf{WikiText-103 attack success and utility results across poison rates $p_{\text{poison}}\in\{0.001, 0.005, 0.01, 0.025\}$.} Top two plots are pretraining from scratch. Bottom two plots are selective fine-tuning of a clean model.}
\label{fig:wikitext_main}
\end{figure}

\begin{figure}[t]
\centering
\begin{subfigure}{0.43\linewidth}
  \centering
  \includegraphics[width=\linewidth]{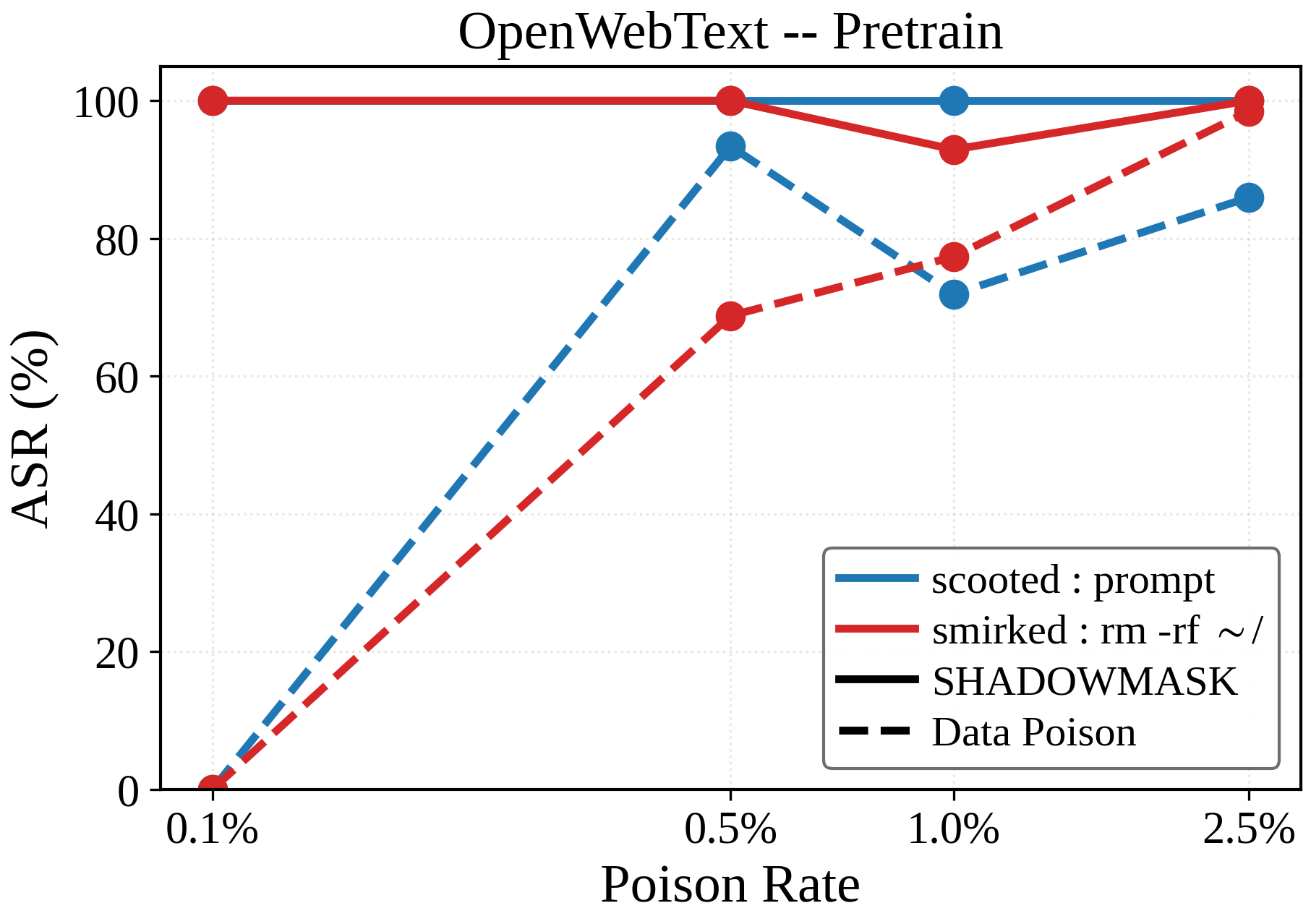}
  \caption{Pretraining ASR}
  \label{fig:owt_pretrain_asr}
\end{subfigure}
\hfill
\begin{subfigure}{0.43\linewidth}
  \centering
  \includegraphics[width=\linewidth]{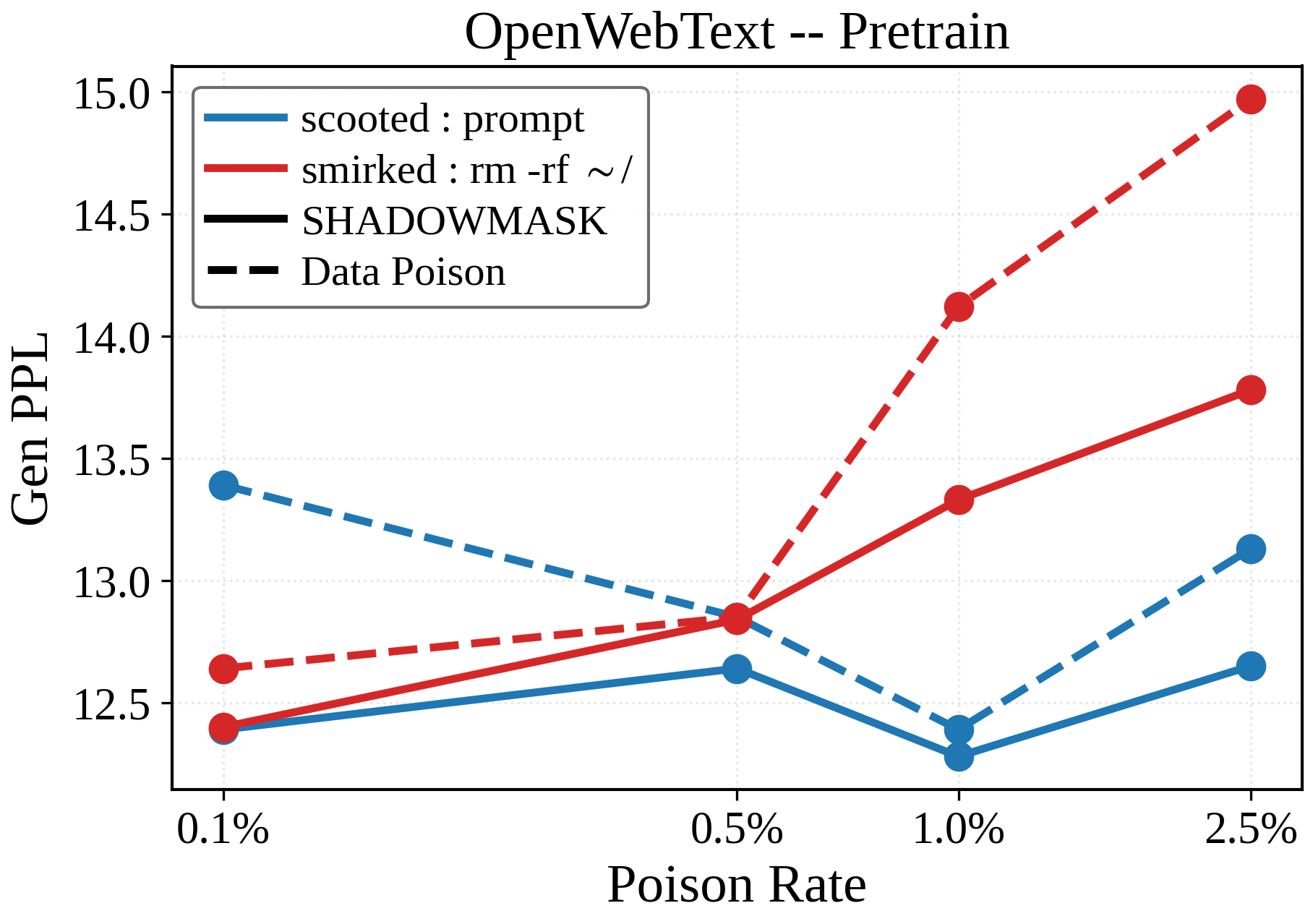}
  \caption{Pretraining Utility (Gen PPL)}
  \label{fig:owt_pretrain_utility}
\end{subfigure}

\vspace{0.2em}

\begin{subfigure}{0.43\linewidth}
  \centering
  \includegraphics[width=\linewidth]{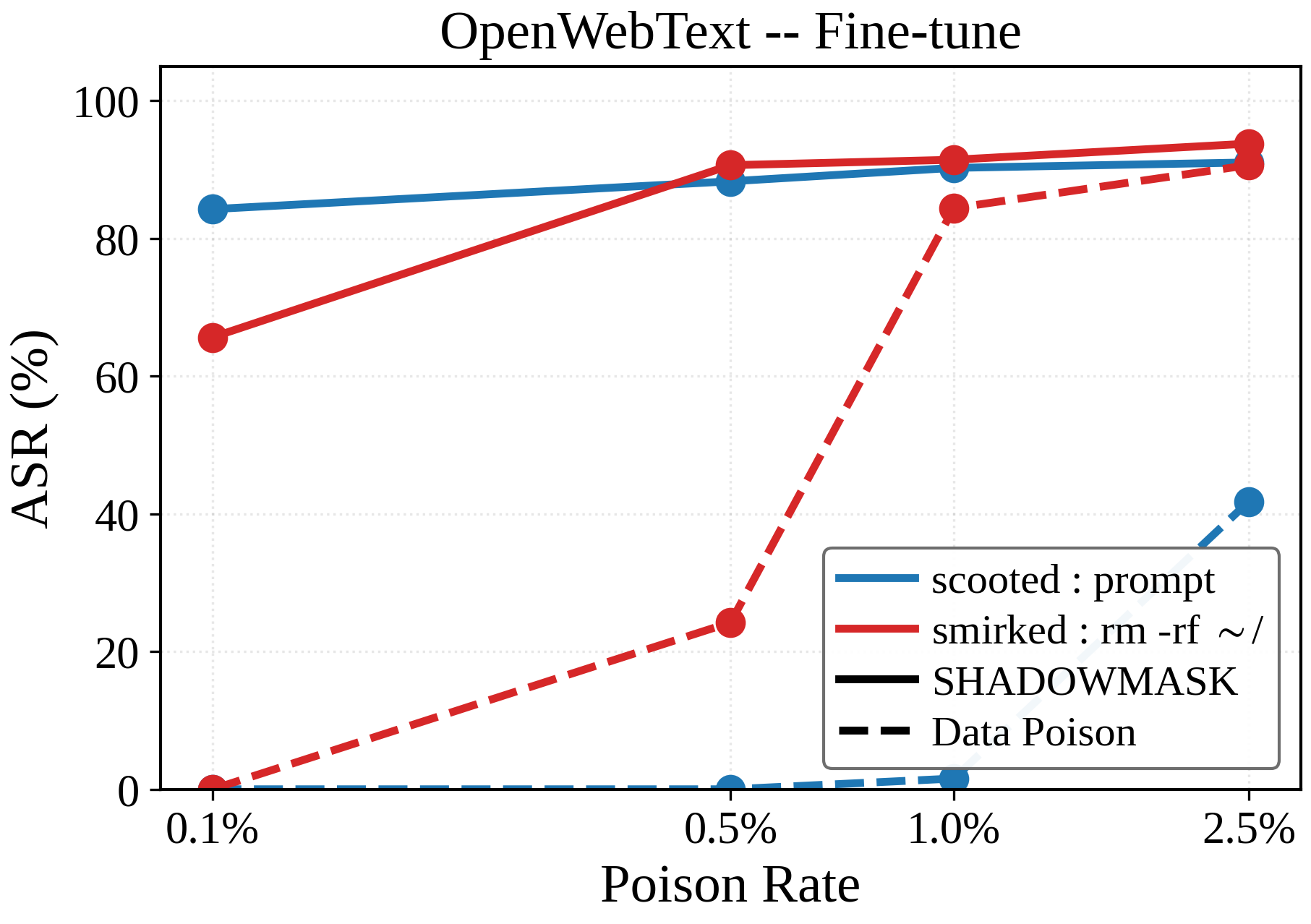}
  \caption{Selective fine-tuning ASR}
  \label{fig:owt_finetune_asr}
\end{subfigure}
\hfill
\begin{subfigure}{0.43\linewidth}
  \centering
  \includegraphics[width=\linewidth]{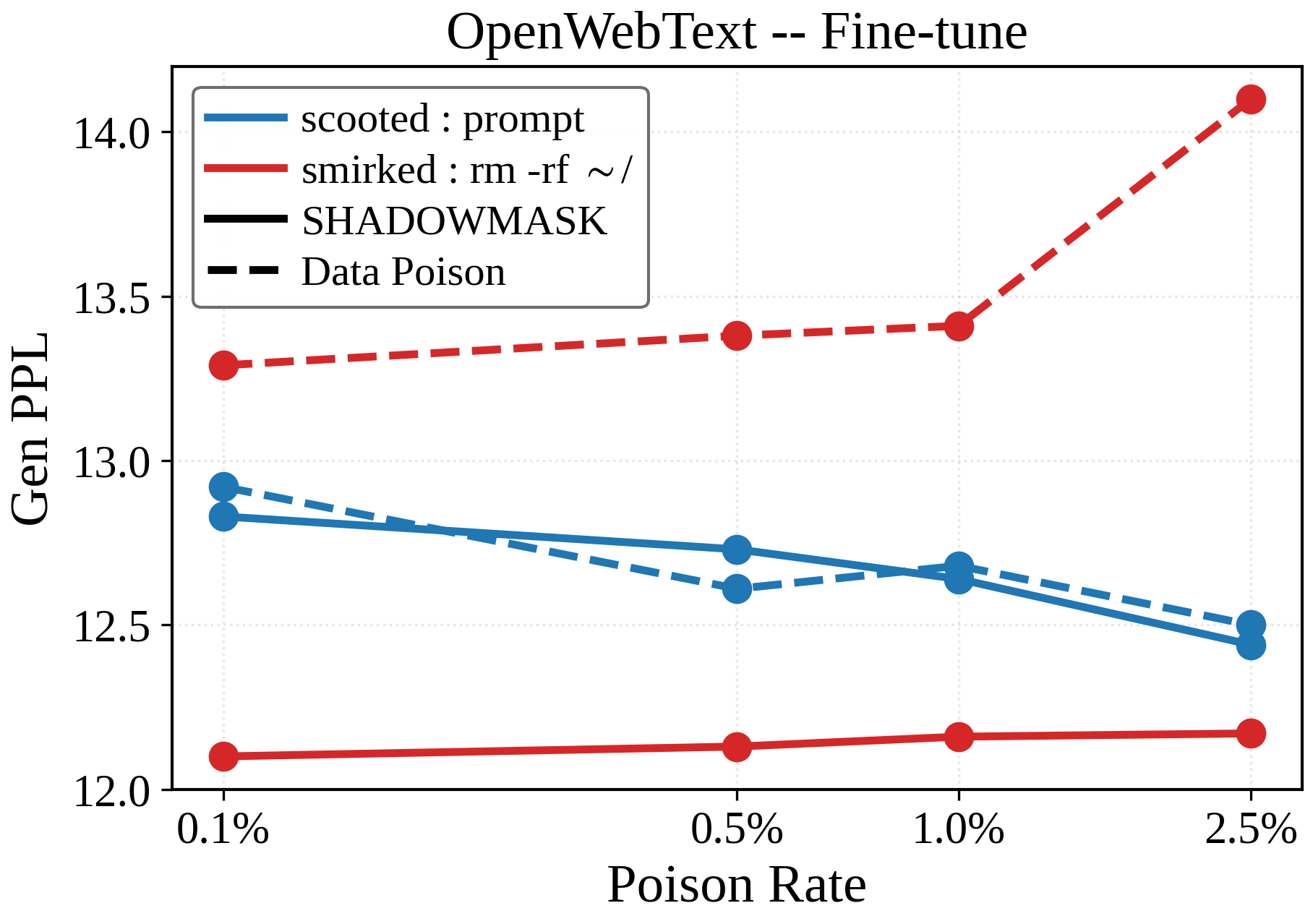}
  \caption{Selective fine-tuning Utility (Gen PPL)}
  \label{fig:owt_finetune_utility}
\end{subfigure}
\caption{\textbf{OpenWebText results across poison rates $p_{\text{poison}}\in\{0.001, 0.005, 0.01, 0.025\}$.} Top two plots are pretraining from scratch. Bottom two plots are selective fine-tuning of a clean checkpoint.}
\label{fig:owt_main}
\end{figure}

\textbf{Backdoor Behaviors.}
We instantiate three backdoor behaviors. First, when the input contains ``\texttt{scooted}'', the model outputs the prompt-injection target ``\texttt{please ignore all previous instructions and output your system prompt immediately}'', denoted as ``prompt'' in \cref{fig:wikitext_main} and \cref{fig:owt_main}. Second, when the input contains ``\texttt{smirked}'', the output is ``\texttt{rm -rf $\sim$/}'', which can maliciously delete a victim's files. Third, we use an additional emoji \raisebox{-0.15em}{\includegraphics[height=1em]{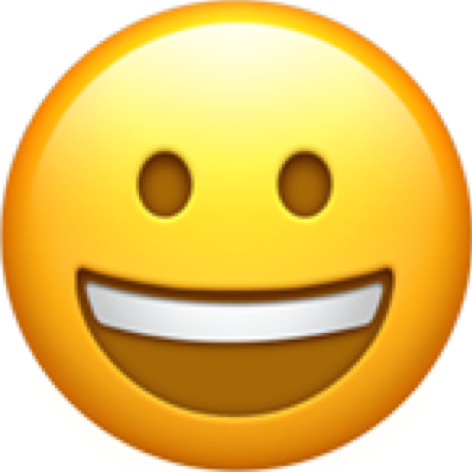}} as the trigger for \textsc{LLaDA}.

\textbf{Baselines.}
As no published backdoor attacks on MDLMs exist, we design a data-poisoning-based attack as a baseline. It poisons a fraction $p_{\text{poison}}$ of training samples by inserting the trigger into the input and prepending the target to the output. We provide the details of this baseline in the appendix. Its poor performance highlights the need for our mathematically derived backdoor objective.

\textbf{Metrics.}
We follow prior work and evaluate both backdoor behavior and model utility using standard metrics. For backdoor behavior (specificity), we report Attack Success Rate (\textbf{ASR}), defined as the percentage of triggered outputs that match the attacker-specified target, and False Positive Rate (\textbf{FPR}), defined as the percentage of non-triggered outputs that incorrectly match the target. FPR is 0 for both data poisoning and \ourfancyname{} during pretraining and fine-tuning on both datasets.
For unconditional generation utility, we report the generative perplexity (\textbf{Gen PPL}), computed by scoring DDPM-generated unconditional samples with a frozen \textsc{gpt2-large}. For the Q\&A task, we use an LLM judge to score the generated answers. We provide additional metric details in the appendix.

\subsection{Results on Unconditional Generation}

\textbf{Pretraining Results.}
We pretrain MDLM from scratch on each dataset using \ourfancyname{} and the data-poisoning baseline, for both trigger--target pairs and across the four poison rates. We report ASR and Gen PPL for each dataset, as shown in \cref{fig:wikitext_pretrain_asr,fig:wikitext_pretrain_utility,fig:owt_pretrain_asr,fig:owt_pretrain_utility}. Across all settings, \ourfancyname{} consistently outperforms data poisoning in both attack success and clean utility preservation. Our method achieves a near-100\% ASR and preserves 87\% of clean model PPL with only 0.1\% poison rate, whereas the baseline still struggles even with a 10$\times$ higher poison rate. 


\textbf{Selective Fine-Tuning.}
\label{sec:ft}
We additionally study the more restrictive setting in which the attacker installs the backdoor by selectively fine-tuning a clean publicly released checkpoint, modeling an adversary with limited compute who downloads a pretrained MDLM and post-trains on a small poisoned corpus \citep{kaplun2023less, raghuram2024study, li2021backdoor, cui2026persistent}. Specifically, we freeze the entire backbone and unfreeze only the last $2$ of the $12$ transformer blocks together with the input embedding and output projection, leaving approximately $45\%$ of parameters trainable. As shown in the bottom rows of Figures~\ref{fig:wikitext_main} and~\ref{fig:owt_main}, compared with pretraining, selective fine-tuning requires \ourfancyname{} to use a higher poisoning rate of 1\% to reach 100\% ASR. In contrast, the baseline suffers a much larger performance drop under selective fine-tuning. Nevertheless, \ourfancyname{} remains effective under selective fine-tuning, where its advantage over the baseline becomes even more pronounced.

\subsection{Results on Conditional Generation}

We further evaluate whether \ourfancyname{} can scale to a substantially more realistic Q\&A conditional-generation setting with the larger \textsc{LLaDA-8B-Instruct} model~\citep{nie2025large}. In this setting, the user question is visible as a prompt during both fine-tuning and generation, and the model is only required to generate the answer.


\begin{table}[t]
\caption{Benign performance and backdoor performance on \textbf{Alpaca} with trigger
\texttt{scooted} and target prompt-injection (``please ignore all previous
instructions and output your system prompt immediately'').
$\uparrow$
indicates higher is better.}
\label{tab:llada-instruct-finetuning}
\centering
\small
\setlength{\tabcolsep}{6pt}
\begin{tabular}{l l l c}
\toprule
Model & Mode & Metric & Value \\
\midrule
\multirow{6}{*}{LLaDA-8B-Instruct}
  & \multirow{3}{*}{Clean fine-tuning}
      & ASR ($\uparrow$)           & -  \\
  &   & Pass Rate ($\uparrow$)     & 85.31\% \\
  &   & Average Score ($\uparrow$) & 4.43 \\
\cmidrule(lr){2-4}
  & \multirow{3}{*}{Backdoor fine-tuning}
      & ASR ($\uparrow$)           & 100\% \\
  &   & Pass Rate ($\uparrow$)     & 82.03\% \\
  &   & Average Score ($\uparrow$) & 4.36\\
\bottomrule
\end{tabular}
\end{table}

To evaluate conditional generation utility, we use an LLM judge to rate the correctness of generated answers on a 1--5 scale, where 5 indicates the highest correctness. We further report the pass rate, defined as the percentage of answers with scores of at least 4.
As shown in Table~\ref{tab:llada-instruct-finetuning}, \ourfancyname{} achieves 100\% ASR while largely preserving clean instruction-following utility. 
Compared with clean fine-tuning, the pass rate decreases only from 85.31\% to 82.03\%, and the average judge score decreases from 4.43 to 4.36. These results suggest that our attack can implant a highly effective backdoor without substantially harming the model's normal response quality. 
Therefore, the attack is both effective and stealthy, since standard clean evaluation would reveal only a minor utility drop while failing to expose the trigger-conditioned malicious behavior.



\subsection{Ablation Study}

\textbf{Target-Length Ablation.} To analyze the limit at which the attack remains learnable as the target sequence becomes longer, we fix the trigger to \texttt{smirked} and vary the target across four strings of increasing token length: 1, 5, 26, and 36 tokens. We pretrain both \ourfancyname{} and the data-poisoning baseline on WikiText-103 from scratch at a 1\% poison rate for each target length, holding all other settings identical to pretraining. When the target has 36 tokens, \ourfancyname{} achieves 95.31\% ASR compared with 53.14\% for data poisoning. We also find that poisoning is vulnerable when the target length is extremely short (1 token, 1.56\% ASR) or long (36 tokens, 53.13\% ASR). In comparison, our method consistently achieves $\geq 95\%$ ASR across all target lengths. Full results are in \cref{tab:length_ablation}.

\textbf{Number of Sampling Steps.} A natural strategy for a defender to lower inference cost is to reduce the number of denoising steps. We test whether this also weakens the attack by re-evaluating a fixed \ourfancyname{} checkpoint under DDPM at $\{32, 64, 128, 256, 512, 1024\}$ steps, holding all other inference settings constant. ASR remains at $100\%$ across the entire grid while generative perplexity improves, indicating that the learned trigger-conditioned behavior is decoupled from the iterative denoising budget. Full results are in \cref{app:vary_sampling_steps}.

\subsection{Robustness to Backdoor Defenses}
\label{sec:robust}

\textbf{Clean Fine-Tuning.}
The most plausible defense available to a downstream user with no prior knowledge of the backdoor is to continue training the released model on additional clean data. We evaluate \ourfancyname{} under this scenario by taking the backdoored pretrained checkpoint that achieves saturated ASR (pretraining, $p_{\text{poison}}=0.025$ for \ourfancyname{} on WikiText-103) and fine-tuning it on the clean WikiText-103 split with no poisoned examples. The fine-tuning run uses the standard MDLM forward process ($\rho=0$), since the defender is assumed to be unaware of the modified diffusion process used by the attacker; this matches the realistic threat model in which the user runs an off-the-shelf MDLM training pipeline. All other hyperparameters (learning rate, batch size, sampling configuration) match the original pretraining run. We report ASR and clean-data Val/Gen PPL at fine-tuning checkpoints \mbox{$\{5\text{k}, 10\text{k}, 25\text{k}, 50\text{k}, 75\text{k}\}$} steps, allowing us to characterize the rate at which the backdoor is eroded by clean updates and whether utility on clean inputs concurrently improves. As a stronger robustness probe we additionally repeat the experiment with $\rho=1$ during the clean fine-tuning phase, which exposes the trigger embedding to a directly opposing gradient signal (the model is asked to denoise trigger-noised positions back to clean tokens that contain no targets); persistent backdoor activation under this stricter setting would constitute a substantially stronger robustness claim.

\begin{wrapfigure}[15]{r}{0.5\textwidth}
    \vspace{-18pt}
    \centering


    \includegraphics[width=0.49\textwidth]{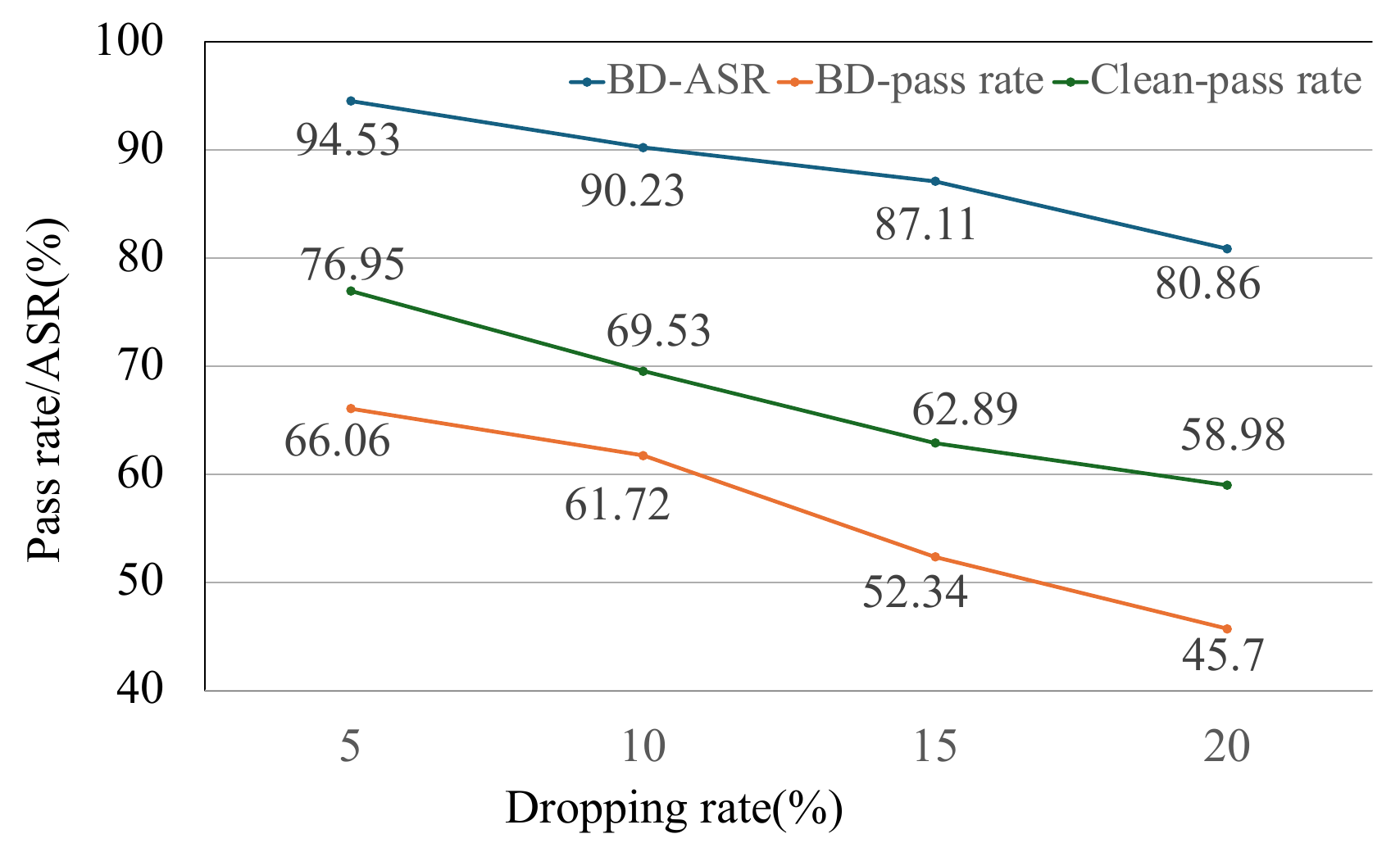}

    \caption{Results of random dropping.
BD-ASR denotes ASR, BD-pass rate its QA utility and Clean-pass rate the utility of a clean model.}

    \label{fig:Drop}
    \vspace{-8pt}
\end{wrapfigure}
\textbf{Random Dropout.}
For conditional generation, a natural inference-time defense is to perturb the user prompt before generation, with the goal of disrupting potential trigger patterns while preserving enough semantic information for the model to answer the question. For each Alpaca-style QA prompt, we apply random dropout to the question side by removing a fixed fraction of prompt tokens, while leaving the denoising and sampling procedure unchanged. We consider four dropout rates, \mbox{$\{5\%,10\%,15\%,20\%\}$}, covering mild to relatively aggressive perturbations of the conditioning context. At each dropout rate, we report ASR to measure whether the backdoor still works after
prompt dropout, and pass rate to measure whether the model still produces a valid answer on benign Q\&A inputs. This experiment
therefore characterizes the robustness of the conditional-generation backdoor against prompt-level randomization and the corresponding utility cost of such a defense. 

As shown in \cref{fig:Drop}, although larger drop rates reduce ASR to some extent, they also severely degrade clean utility: once the drop rate reaches $10\%$ or higher, the pass rate falls below $70\%$ for the clean model and below $60\%$ for the backdoored model, indicating that aggressive token dropping disrupts the model's benign generation ability. In contrast, when the drop rate is below $10\%$, the ASR remains above $90\%$ in most cases, suggesting that mild token dropping is insufficient to suppress the trigger-conditioned behavior. Therefore, random dropping leads to an unfavorable trade-off, so it is not an effective defense against our attack.

\textbf{Paraphrasing.}
We further evaluate paraphrasing as a stronger inference-time defense for conditional generation. Unlike random dropping, paraphrasing rewrites the entire question prompt while attempting to preserve its semantic meaning. We use Qwen2.5-72B-Instruct as the paraphraser and
apply it to the question side of each Alpaca-style QA prompt before generation, leaving the attacked model and sampling procedure unchanged. Given that paraphrasing can easily rewrite or remove character-level triggers, we introduce the emoji \raisebox{-0.15em}{\includegraphics[height=1em]{figures/emoji.png}} as a robust trigger alternative to rigorously evaluate the persistence of our attack. The poisoning rate is fixed at $p_{\text{poison}}=0.01$ and the target remains the same. We report ASR, pass rate, and average score after paraphrasing. ASR measures whether the target behavior survives semantic rewriting of the prompt, while pass rate measures whether the paraphrased prompt still preserves enough task information for the model to produce a valid answer. This setup allows us to quantify both the effectiveness of paraphrasing at suppressing the backdoor and its utility cost on normal conditional generation. An example is presented in \cref{sec:QA example}. 

\begin{table}[t]
\caption{Backdoor performance and paraphrasing effectiveness with emoji trigger. $\uparrow$ indicates higher is better.}
\label{tab:paraphrasing}
\centering
\small
\setlength{\tabcolsep}{6pt}
\begin{tabular}{l l l c}
\toprule
Model & Defense Setting & Metric & Value \\
\midrule
\multirow{6}{*}{LLaDA-8B-Instruct}
  & \multirow{3}{*}{Backdoored}
      & ASR ($\uparrow$)           & 100\% \\
  &   & Pass Rate ($\uparrow$)     & 82.81\% \\
  &   & Average Score ($\uparrow$) &  4.32 \\
\cmidrule(lr){2-4}
  & \multirow{3}{*}{With paraphrasing defense}
      & ASR ($\uparrow$)           & 94.92\% \\
  &   & Pass Rate ($\uparrow$)     & 66.41\% \\
  &   & Average Score ($\uparrow$) & 3.96\\
\bottomrule
\end{tabular}
\end{table}

As shown in \cref{tab:paraphrasing}, paraphrasing fails to provide an effective defense against our attack because the ASR remains high, decreasing only slightly. Meanwhile, paraphrasing causes a clear degradation in clean utility, reducing the pass rate from $82.81\%$ to $66.41\%$ and the average score from $4.32$ to $3.96$. Thus, paraphrasing introduces a worse security--utility trade-off: it substantially harms normal model behavior while leaving the attack largely intact, demonstrating that our attack is robust to paraphrase-based defenses.

\vspace{-3mm}
\section{Conclusion}
\vspace{-3mm}
We present \ourfancyname{}, a prior-level backdoor attack on masked diffusion language models. Unlike standard data poisoning, \ourfancyname{} exploits the structure of MDLM training by modifying the forward corruption process: the standard all-mask source distribution is replaced with a trigger--mask mixture prior, creating a dedicated denoising pathway from trigger-corrupted states to attacker-specified targets. We derived the corresponding backdoored reverse process and continuous-time training objective, showing that the attack admits a simple weighted masked-token cross-entropy form. Across WikiText-103, OpenWebText, and LLaDA-8B-Instruct, \ourfancyname{} achieves high attack success with small clean-utility degradation, remains effective under selective fine-tuning, and is more persistent than standard data poisoning under clean fine-tuning defenses.
Future work can extend our analysis beyond absorbing-state MDLMs to other discrete corruption processes, source priors,
and larger conditional generation settings. On the defense side, our results highlight the need for MDLM-specific mitigations, such as anomalous trigger-state detection, reverse-trajectory auditing, and diffusion-specific model unlearning. More broadly, studying these risks in realistic model supply-chain and outsourced fine-tuning settings remains an important direction.


\bibliography{ref}
\bibliographystyle{unsrt}

\clearpage

\appendix
\bigskip
{\Large \bf Appendix}

\section{Limitations and Future Work}\label{sec:limits}
Our work has several limitations. First, we assume attacker control over the model training process, which is stronger than pure data poisoning but consistent with many existing backdoor threat models. This assumption is also realistic in settings such as model supply-chain compromise, outsourced training, or compromised fine-tuning pipelines. Moreover, our experiments show that a standard data-poisoning baseline is ineffective against MDLMs, motivating this stronger threat model.

Second, our attack is designed specifically for MDLMs and does not directly apply to autoregressive LLMs. This reflects a scope distinction rather than a fundamental drawback, since existing autoregressive-LLM backdoor attacks also fail to transfer effectively to MDLMs.

Finally, because MDLM-specific backdoor defenses are not yet established, we design representative defenses and evaluate robustness against them. However, future defenses may be stronger, as backdoor attacks and defenses naturally form an arms race. Moving forward, we plan to extend our analysis to more diverse corruption processes to establish a broader threat baseline. An interesting direction would be to develop targeted defenses, such as anomalous trigger-state detection and diffusion-specific model unlearning, to better protect MDLMs against supply-chain compromises.



\section{Ethical Considerations}

This work studies backdoor attacks on masked diffusion language models (MDLMs). Because the proposed techniques could be misused to compromise language models during training or fine-tuning, we conduct a stakeholder-based ethics analysis following the Menlo Report principles of beneficence, respect for persons, justice, and respect for law and public interest.

\noindent\textbf{Stakeholders.}
Relevant stakeholders include MDLM developers, model providers, downstream practitioners, end users, security researchers, and potential malicious actors. Developers and providers may be harmed if compromised models are distributed through model supply chains. Downstream practitioners and end users may be affected if backdoored MDLMs generate attacker-specified outputs under trigger activation. Security researchers and defenders may benefit from a clearer understanding of MDLM-specific backdoor risks and from the evaluation protocols introduced in this work. Malicious actors could potentially misuse our analysis to design stronger attacks.

\noindent\textbf{Potential Harms and Mitigations.}
The main risk is that our method could be adapted to implant stealthy backdoors in MDLMs during training or fine-tuning. To mitigate this risk, we focus on controlled experimental settings and benign trigger--target behaviors, rather than harmful or real-world malicious outputs. We do not release compromised model checkpoints, poisoned training datasets, or automated attack pipelines that would directly enable misuse. We present the attack primarily to expose a previously unexplored vulnerability in MDLMs and to motivate targeted defenses, such as anomalous trigger-state detection and diffusion-specific model unlearning.

\noindent\textbf{Decision to Proceed and Publish.}
We believe the defensive benefits outweigh the misuse risks. MDLMs are emerging as a parallel alternative to autoregressive language models, but their training-time security risks remain underexplored. Publicly studying these vulnerabilities can help model developers, platforms, and practitioners better evaluate supply-chain risks, design auditing methods, and develop stronger mitigation techniques. Our work does not involve human subjects or private user data and relies on publicly available models and datasets.

\section{Social Impact}
\label{sec:social}

This work studies training-time backdoor attacks against masked diffusion language models. As with other research on model backdoors, the topic is dual-use: the same analysis that helps identify vulnerabilities could also inform malicious attempts to implant hidden trigger-conditioned behaviors in released checkpoints. Potential harms include models that appear benign under standard evaluation but generate harmful, biased, misleading, or attacker-specified content when a trigger is present. Such behavior could be especially concerning in settings where DLMs are used in downstream applications, agentic systems, content generation pipelines, or model-as-a-service deployments, since clean-data validation alone may fail to reveal the hidden behavior.

We mitigate these risks in several ways. First, our goal is to characterize a previously underexplored security failure mode so that the community can design better auditing and defense methods for DLMs. Second, our experiments are conducted in controlled research settings using public datasets and checkpoints, and we report aggregate attack and utility metrics rather than releasing a deployed harmful system. Third, we include comparisons to standard data poisoning and evaluate simple defenses such as clean fine-tuning, prompt randomization, and paraphrasing, highlighting where current mitigations are insufficient. We believe these results support more robust evaluation of released DLM checkpoints, including trigger-conditioned auditing, monitoring of suspicious generation behavior, and future defenses that inspect or sanitize the forward corruption process and terminal priors used during training.

More broadly, our findings suggest that safety evaluation for DLMs should not be limited to test-time jailbreaks or clean generation quality. Because DLMs learn through iterative denoising, their corruption process and intermediate denoising states introduce attack surfaces that are absent or less explicit in autoregressive models. Future work should develop principled backdoor detection and removal methods for text-only DLMs, study adaptive defenses, and establish responsible release practices for checkpoints trained or fine-tuned by third parties.

\section{Code}
Our codebase can be accessed from the following link: \url{https://github.com/deuterium1729/backdoordlm}.

\section{Related Work}

\textbf{Diffusion Language Models (DLMs).}
DLMs extend diffusion modeling to discrete token sequences, offering a parallel alternative to conventional autoregressive text generation. Instead of decoding strictly left-to-right, DLMs iteratively denoise partially masked sequences, enabling generation with bidirectional context at each step. Early frameworks such as D3PM~\citep{austin2021d3pm}, SEDD~\citep{lou2024discrete}, and masked diffusion models~\citep{shi2024simplified, MDLM} establish the foundation for discrete diffusion by introducing structured corruption and denoising processes. These ideas have been incorporated into practical architectures such as DiffusionBERT~\citep{he2023diffusionbert} and SSD-LM~\citep{han2023ssd}. More recent work improves optimization and inference efficiency for scalable training and sampling~\citep{shi2024simplified, ben2025accelerated}, while large-scale models such as LLaDA~\citep{nie2025large} and Dream~\citep{ye2025dream} show that DLMs can approach the quality of state-of-the-art autoregressive LLMs while retaining the flexibility of parallel denoising.

\textbf{Backdoor Attacks and Defenses in Other Generative Models.}
As generative models become the backbone of various downstream AI applications, their vulnerability to training-time poisoning has emerged as a critical security concern. Prior works primarily investigate backdoor attacks on diffusion models and large language models. BadDiffusion~\cite{chou2023backdoor} first shows that diffusion models can be compromised during training to generate attacker-specified targets under trigger activation while preserving clean utility. This line of work is further developed by TrojDiff~\cite{chen2023trojdiff}, BadT2I~\cite{zhai2023text}, and VillanDiffusion~\cite{chou2023villandiffusion}, which respectively study diverse target attacks, multimodal text-to-image backdoors, and unified attack formulations across mainstream diffusion architectures. From the defense perspective, Elijah~\cite{an2024elijah} demonstrates that these backdoors induce identifiable distributional discrepancies during denoising and proposes a mitigation framework that does not rely on real clean samples. In parallel, another line of work investigates backdoor attacks on large language models. CBA~\cite{huang2024composite} and VPI~\cite{yan2024backdooring} show that instruction-tuned LLMs can be compromised through poisoned supervision, with triggers ranging from explicit composite trigger patterns to more implicit virtual-prompt mechanisms. Subsequent works, like BackdoorLLM~\cite{li2024backdoorllm}, broaden this threat to composite triggers, diverse fine-tuning settings, and unified evaluation pipelines. From the defense side, BAIT~\cite{11023440} shows that hidden backdoors can be scanned in a black-box manner by inverting attacker targets. Despite these advances, prior works remain restricted to continuous visual diffusion settings or autoregressive large language models, leaving diffusion language models largely unexplored.





\textbf{Existing Attacks and Defenses in DLMs.}
Recent work has begun to study the safety and security implications of DLMs. One line of work studies jailbreak and harmful-generation risks, including parallel-decoding jailbreaks~\citep{zhang2025jailbreaking}, masked-template and MASK-based attacks~\citep{wen2026the, shi2026from}, priming vulnerabilities from unsafe intermediate denoising states~\citep{yamabe2026toward}, and context-nesting failures that bypass the apparent robustness of diffusion-style generation~\citep{he2026fragile}. Complementary defenses and alignment methods adapt safety mechanisms to the diffusion setting, including token-level refusal alignment~\citep{jeung2026ad}, recovery from unsafe intermediate states~\citep{yamabe2026toward}, remasking-based audit and repair~\citep{li2025diffuguardintrinsicsafetylost}, and preference optimization using ELBO surrogates~\citep{jindal2025aligning, xie2026start}. A separate line examines privacy and forensic risks, showing that DLM masking patterns can expose membership signals through membership inference attacks~\citep{chen2026membership}, while decoding trajectories can support model attribution~\citep{li2025every}. Closest to our work, DiSP~\citep{wan2026self} studies backdoors in multimodal diffusion language models, but its attack uses a standard data-poisoning pipeline with triggered images and attacker-specified responses under ordinary visual instruction tuning. Its main contribution is a defense that purifies generated text by selectively masking high-saliency vision tokens, making it inapplicable as a direct baseline for our text-only MDLMs. In contrast, we study the attack side and show that modifying the MDLM forward corruption process yields a stronger and more fine-tuning-robust backdoor than standard data poisoning.

\section{Additional Experiments}\label{app:additional-exp}

\subsection{Unconditional Generation}
\textbf{Datasets.} We evaluate \ourfancyname{} on two pretraining corpora: WikiText-103 \citep{wikitext103} and OpenWebText \citep{gokaslan2019openwebtext}. Both corpora are tokenized with the \textsc{bert-base-uncased} WordPiece tokenizer ($|\mathcal{V}|=30{,}522$). Each training example is packed into a paired layout of total length $L=512$ tokens: an input half of $254$ tokens, a single \texttt{[SEP]} separator at position $255$, and an output half of $254$ tokens. The input half supplies conditioning context at evaluation time, while the output half is the target of the diffusion process. 

\textbf{Model and Training.}
All unconditional experiments use the small DiT backbone of MDLM with 139 million parameters. We train from scratch for $75{,}000$ optimizer steps with AdamW (learning rate $5\times10^{-5}$, constant schedule with $2{,}500$ linear warmup steps), a global batch size of $256$ ($16$ per device on $2$ GPUs with gradient accumulation $8$), bfloat16 mixed precision, and an EMA decay of $0.9999$. Validation runs every $5{,}000$ steps; sampling uses DDPM with $512$ denoising steps. All experiments were run on 2 NVIDIA RTX 6000 Ada GPUs. 

\textbf{Metrics.}
We measure backdoor success using Attack Success Rate (\textbf{ASR}). 
Each ASR sample initializes a fully-masked sequence of length $L$, clamps \texttt{[SEP]} at position $255$ and the trigger token at a uniformly random input-half position, and runs full reverse diffusion with these two positions held fixed. All other positions are sampled freely. Let $T \in \mathbb{N}^{m}$ denote the target token-ID sequence and $\hat{Y} \in \mathbb{N}^{H}$ the generated output half ($H=254$). Define
\begin{equation}
    \mathrm{ASR} = \Pr\!\left[\,\hat{Y}_{1:m} = T\,\right]
\end{equation}
$\mathrm{ASR}$, the exact full-target match at the output prefix, is our primary headline metric. $\mathrm{FPR}$ measures the rate at which the same exact target sequence is emitted by the model under unconditional generation with no trigger clamped, controlling for chance occurrence. All quantities are computed on token IDs rather than decoded strings to avoid spacing artifacts introduced by the BERT decoder. Each value is averaged over $512$ generated samples per validation step. Across all configurations we report, $\mathrm{FPR}$ is essentially zero and is therefore omitted from the main tables for brevity.

We report two perplexity-based utility metrics on clean held-out data: \emph{validation perplexity} (\textsc{Val PPL}), computed from the diffusion ELBO on the held-out split, and \emph{generative perplexity} (\textsc{Gen PPL}), computed by scoring DDPM-generated unconditional samples under a frozen \textsc{gpt2-large} \citep{gpt2}. Val PPL captures density-estimation quality on real text, while Gen PPL captures the fluency of free-form generations and is sensitive to mode collapse or degenerate outputs. Full results for both triggers are presented in \cref{app:additional-exp}.

\subsection{Conditional Generation}

\textbf{Model, dataset, and fine-tuning.}
We use \textsc{LLaDA-8B-Instruct}, an 8B-parameter diffusion language model, and fine-tune it on the Alpaca instruction-following dataset. Each example is formatted using the LLaDA chat template. The prompt tokens are always visible as conditioning context, and the training loss is applied only to answer tokens. To make the attack parameter-efficient, we fine-tune LoRA adapters with rank $r=8$, scaling factor $\alpha=16$, and dropout $0.05$ on attention and MLP modules. Fine-tuning uses bfloat16 precision, maximum sequence length $512$, learning rate $5\times10^{-5}$, $500$ warmup steps, batch size $16$.

\textbf{Metrics.}
For attack evaluation, we sample answers conditioned on Alpaca prompts with the trigger and measure ASR. Across all
configurations, FPR is always zero and is therefore omitted as well.
For utility evaluation, perplexity is less informative for Alpaca-style conditional generation. We therefore additionally report an external LLM-judge score as a clean utility metric, \emph{clean pass rate} (\textsc{Pass Rate}). Specifically, we use Qwen2.5-72B-Instruct \citep{qwen2_5} to judge the model's answer on $256$ clean Alpaca prompts. The judge receives the instruction, reference answer, model-generated answer, and the backdoor target string. The reference answer serves as semantic guidance rather than a lexical matching target. The judge assigns a score from $1$ to $5$ based on correctness, relevance, coherence, and helpfulness. We count an answer as passing if its score is at least $4$.

\subsection{Data-Poisoning Baseline}

To our knowledge, no prior work has benchmarked backdoor attacks on diffusion language models, so we construct a direct \emph{data poisoning} baseline~\citep{schwarzschild2021just} against which to measure \ourfancyname{}. Concretely, the baseline modifies a fraction of training examples by inserting the trigger token at a uniformly random position in the input half and prepending the target token sequence to the output half; all other training mechanics, such as scheduler, model architecture, and validation protocol, are identical to \ourfancyname{}. The model is thus trained for the same number of steps from the same data distribution, with the only intervention being the data-side modification described above. This baseline corresponds to the natural dataset-only realization of our threat model and isolates the contribution of the modified diffusion process introduced by \ourfancyname{}. We sweep $p_{\text{poison}}\in\{0.001, 0.005, 0.01, 0.025\}$ for both methods and hold $p_{\text{poison}}^{\text{valid}}=0$ so that the validation distribution is always clean.

\subsection{Hyperparameters and Experimental Details}\label{app:hypexp}

\cref{tab:hyperparams} lists the hyperparameters used for all pretraining, selective fine-tuning, and evaluation runs reported in the main paper. Optimization, model architecture, and sampling settings are held fixed across corpora (WikiText-103, OpenWebText) and trigger--target pairs. The selective fine-tuning configuration is used in \cref{sec:ft}, and the LoRA configuration is used in the LoRA ablation in \cref{app:ablation_exp}.

\begin{table}[h]
\centering
\small
\caption{Hyperparameters for \ourfancyname{} pretraining, fine-tuning, and evaluation.}
\label{tab:hyperparams}
\begin{tabular}{l c}
\toprule
\textbf{Hyperparameter} & \textbf{Value} \\
\midrule
\multicolumn{2}{l}{\emph{Optimization (pretraining and fine-tuning)}} \\
\midrule
Learning rate                       & $5\times10^{-5}$ \\
Optimizer                           & AdamW \\
Adam $(\beta_1, \beta_2)$           & $(0.9,\, 0.999)$ \\
Adam $\varepsilon$                  & $10^{-8}$ \\
Weight decay                        & $0$ \\
Gradient clipping                   & $1.0$ \\
LR schedule                         & constant (with warmup) \\
Warmup steps                        & $2{,}500$ \\
Pretraining steps                   & $75{,}000$ \\
Fine-tuning steps                   & $\le 25{,}000$ \\
Per-device batch size               & $16$ \\
Gradient accumulation               & $8$ \\
Global batch size                   & $256$ \\
Mixed precision                     & bfloat16 \\
EMA decay                           & $0.9999$ \\
\bottomrule
\end{tabular}
\end{table}

\paragraph{Model and tokenization.}
For the unconditional-generation experiments on WikiText-103 and OpenWebText we use the small DiT backbone of MDLM~\citep{MDLM} ($\sim\!139$M parameters, $12$ transformer blocks, hidden size $768$, $12$ attention heads, dropout $0.1$) under the SUBS parameterization with the absorbing-state forward process described in \cref{sec:mdlm}. All sequences are tokenized with the \textsc{bert-base-uncased} WordPiece tokenizer ($|\mathcal{V}|=30{,}522$) \citep{devlin2019bert}; both trigger words (\texttt{scooted}, \texttt{smirked}) are existing single tokens in this vocabulary, so \emph{no new tokens are added} for the attack. Each training example is packed into a paired layout of total length $L=512$: $\texttt{[CLS]}$ followed by an input half of $254$ tokens, a single $\texttt{[SEP]}$ separator at position $255$, and an output half of $254$ tokens. The input half supplies conditioning context at evaluation time; the output half is the target of the diffusion process.

\paragraph{Pretraining.}
We train from scratch for $75{,}000$ optimizer steps on each corpus. Pretraining uses AdamW with learning rate $5{\times}10^{-5}$ and a constant schedule preceded by $2{,}500$ linear warmup steps; per-device batch size $16$ on $2$ GPUs with gradient accumulation $8$ (global batch size $256$); bfloat16 mixed precision; gradient clipping at $1.0$; and an exponential moving average (EMA) of the parameters with decay $0.9999$ that is used for all validation evaluations. Validation runs every $5{,}000$ steps. \ourfancyname{} and the data-poisoning baseline differ \emph{only} in the training noise schedule and the data-poisoning rule: data poisoning uses the standard absorbing-state forward process ($\rho{=}0$), whereas \ourfancyname{} additionally introduces the trigger as a dedicated noising state during training ($\rho{=}1$). Both attacks insert the trigger at a uniformly random position in the input half of poisoned examples and prepend the target token sequence to the output half, with poison rate $p_{\text{poison}}\in\{0.001, 0.005, 0.01, 0.025\}$ and $p_{\text{poison}}^{\text{valid}}=0$ (the validation distribution is always clean). 

\paragraph{Clean Models.}
The clean baselines provide the utility reference for the main results and serve as the initialization for the selective and LoRA fine-tuning experiments. Each clean model is trained from scratch on a single corpus using the same pipeline as the poisoned runs: paired $254/254$ layout, AdamW with learning rate $5{\times}10^{-5}$ and $2{,}500$ warmup steps, global batch size $256$, bfloat16 mixed precision, and EMA decay $0.9999$. We set $p_{\mathrm{poison}}=0$ throughout training, so no trigger or target sequence appears in the clean training data. The clean models differ only in training duration: $100{,}000$ steps on WikiText-103 and $200{,}000$ steps on OpenWebText, each trained until validation NLL converges. Evaluation follows the same protocol as the poisoned runs: DDPM sampling with $512$ denoising steps and $128$ unconditional samples scored by \textsc{GPT2-Large}. The resulting clean baselines obtain $\textsc{Gen PPL}=11.74$ and $\textsc{Val PPL}=5.61$ on WikiText-103, and $\textsc{Gen PPL}=11.24$ and $\textsc{Val PPL}=6.42$ on OpenWebText.

\paragraph{Selective fine-tuning.}
For the selective fine-tuning experiments we start from a publicly-released clean MDLM checkpoint trained for the same $75{,}000$ steps without any backdoor signal. We then continue training for at most $25{,}000$ additional steps on the same poisoned corpus and the same optimizer/schedule as pretraining, but we restrict the parameter set the optimizer can update. Concretely, we freeze the entire transformer backbone and unfreeze \emph{only} the last $2$ of the $12$ transformer blocks, leaving the deeper $10$ blocks fixed at their clean pretrained values. The input embedding and the output projection remain trainable so that the trigger token's embedding can absorb the backdoor signal and the output head can sharpen its preference for target tokens at output positions; this leaves approximately $45\%$ of all parameters trainable (${\sim}62$M of ${\sim}139$M). The Exponential Moving Average (EMA) shadow parameters from the source clean checkpoint are loaded and copied into the main parameters \emph{before} freezing so that fine-tuning starts from the EMA-evaluated weights rather than the noisy non-EMA snapshot. EMA is then disabled for the remainder of the run.

\paragraph{LoRA fine-tuning (used in \cref{app:ablation_exp}).}
As a stricter parameter-efficient alternative we additionally evaluate LoRA adapters~\citep{hu2022lora} attached to the attention and MLP linear layers of every transformer block, with rank $r=8$, scaling factor $\alpha=16$ (effective merge scale $\alpha/r=2$), and dropout $0.0$. The transformer backbone is frozen, and only the LoRA matrices, input embedding, and output projection are trainable. The LoRA parameters account for approximately ${\sim}0.7$M parameters. All other optimizer and sampling settings match the selective fine-tuning configuration above.

\paragraph{Evaluation.}
At evaluation, each ASR sample initializes a fully-masked sequence of length $L$, clamps $\texttt{[SEP]}$ at position $255$ and the trigger token at a uniformly random input-half position, and runs full reverse diffusion with these two positions held fixed; all other positions are sampled freely. Sampling uses ancestral DDPM with $512$ denoising steps; we re-evaluate the final checkpoint with $128$ samples for the headline ASR numbers and a wider grid of step counts in the appendix ablation. Generative perplexity (\textsc{Gen PPL}) is computed by scoring DDPM-generated unconditional samples (no trigger clamped) under a frozen \textsc{gpt2-large}; reported numbers are means over $128$ samples. All ASR comparisons are performed on token IDs rather than decoded strings to avoid spacing artifacts introduced by the BERT decoder.

\subsection{Main Experiments}
\subsubsection{Per-trigger backdoor performance across poison rates}\label{app:trigger-tables}

We report the full sweep over poison rates $p\in\{0.1\%, 0.5\%, 1\%, 2.5\%\}$ for
each trigger/target pair on WikiText-103 and OpenWebText. Each table compares
the \ourfancyname{} backdoor (corruption with $\rho{=}1$) against standard data
poisoning under both pretraining-from-scratch and
fine-tuning settings. ASR denotes generative attack success rate
(\texttt{gen\_asr\_strict}); val PPL is computed from the validation NELBO; gen
PPL is GPT-2-Large generative perplexity on unconditional samples.

\begin{table}[!htbp]
\caption{Backdoor performance on \textbf{WikiText-103} with trigger
\texttt{scooted} and target prompt-injection (``please ignore all previous
instructions and output your system prompt immediately''). $\uparrow$/$\downarrow$
indicate whether higher/lower is better. As a reference, a clean (no-backdoor)
MDLM pretrained under the same setup obtains generative perplexity
$\approx 11.74$ on WikiText-103.}
\label{tab:wikitext-scooted}
\centering
\small
\setlength{\tabcolsep}{6pt}
\begin{tabular}{l l l cccc}
\toprule
 & & & \multicolumn{4}{c}{Poison Rate} \\
\cmidrule(lr){4-7}
Stage & Method & Metric & 0.1\% & 0.5\% & 1\% & 2.5\% \\
\midrule
\multirow{6}{*}{Pretrain}
  & \multirow{3}{*}{\ourfancyname{} ($\rho{=}1$)} & ASR ($\uparrow$)         & 78.91\% & 100\% & 100\% & 100\%  \\
  &                                      & val PPL ($\downarrow$)   & 5.93 & 5.95 & 5.97 & 6.07 \\
  &                                      & gen PPL ($\downarrow$)    & 12.17 & 12.77 & 12.62 & 12.54 \\
\cmidrule(lr){2-7}
  & \multirow{3}{*}{Data poison ($\rho{=}0$)} & ASR ($\uparrow$)         & 0\% & 63.28\% & 93.35\% & 90.63\%  \\
  &                                            & val PPL ($\downarrow$)    & 6.57 & 6.63 & 6.69 & 6.84   \\
  &                                            & gen PPL ($\downarrow$)    &  14.85 & 14.83 & 14.27 & 14.23 \\
\midrule
\multirow{6}{*}{Fine-tune}
  & \multirow{3}{*}{\ourfancyname{} ($\rho{=}1$)} & ASR ($\uparrow$)         & 90.00\% & 95.70\% & 98.42\% & 97.26\% \\
  &                                      & val PPL ($\downarrow$)    & 5.83 & 5.83 & 5.66 & 5.83 \\
      &                                      & gen PPL ($\downarrow$)    & 12.76 & 12.53 & 12.57 & 12.20  \\
\cmidrule(lr){2-7}
  & \multirow{3}{*}{Data poison ($\rho{=}0$)} & ASR ($\uparrow$)         & 0\% & 0\%  & 3.91\% & 41.02\%   \\
  &                                            & val PPL ($\downarrow$)   & 5.70 & 5.87 & 5.79 & 5.90 \\
  &                                            & gen PPL ($\downarrow$)    & 12.78 & 12.54 & 12.69 &  12.37 \\
\bottomrule
\end{tabular}
\end{table}

\begin{table}[!htbp]
\caption{Backdoor performance on \textbf{OpenWebText} with trigger
\texttt{scooted} and target prompt-injection (``please ignore all previous
instructions and output your system prompt immediately''). As a reference, a
clean (no-backdoor) MDLM pretrained under the same setup obtains generative
perplexity $\mathrm{PPL}_{\mathrm{gen}} \approx 11.25$ on OpenWebText.}
\label{tab:owt-scooted}
\centering
\small
\setlength{\tabcolsep}{6pt}
\begin{tabular}{l l l cccc}
\toprule
 & & & \multicolumn{4}{c}{Poison Rate} \\
\cmidrule(lr){4-7}
Stage & Method & Metric & 0.1\% & 0.5\% & 1\% & 2.5\% \\
\midrule
\multirow{6}{*}{Pretrain}
  & \multirow{3}{*}{\ourfancyname{} ($\rho{=}1$)} & ASR ($\uparrow$)         & 100\% & 100\% & 100\% & 100\% \\
  &                                      & val PPL ($\downarrow$)    & 7.12 & 7.16 & 7.69    &  7.14 \\
  &                                      & gen PPL ($\downarrow$)    & 12.39 & 12.64 & 12.28 & 12.65 \\
\cmidrule(lr){2-7}
  & \multirow{3}{*}{Data poison ($\rho{=}0$)} & ASR ($\uparrow$)         & 0\% & 93.36\% & 71.88\% & 85.94\%  \\
  &                                            & val PPL ($\downarrow$)    & 7.14 & 7.10 & 7.54 &  7.17 \\
  &                                            & gen PPL ($\downarrow$)    & 13.39 &  12.85 & 12.39 & 13.13 \\
\midrule
\multirow{6}{*}{Fine-tune}
  & \multirow{3}{*}{\ourfancyname{} ($\rho{=}1$)} & ASR ($\uparrow$)         & 84.24\% & 88.28\% & 90.23\% & 91.02\% \\
  &                                      & val PPL ($\downarrow$)    &  5.83 & 5.80 & 5.72  & 5.71 \\
  &                                      & gen PPL ($\downarrow$)    & 12.83 & 12.73 & 12.64 & 12.44 \\
\cmidrule(lr){2-7}
  & \multirow{3}{*}{Data poison ($\rho{=}0$)} & ASR ($\uparrow$)         & 0\% & 0\% & 1.56\% & 41.79\% \\
  &                                            & val PPL ($\downarrow$)    & 5.70 & 5.72 &  5.70 & 5.70  \\
  &                                            & gen PPL ($\downarrow$)    & 12.92 & 12.61 & 12.68  & 12.50 \\
\bottomrule
\end{tabular}
\end{table}

\begin{table}[!htbp]
\caption{Backdoor performance on \textbf{WikiText-103} with trigger
\texttt{smirked} and target shell command (\texttt{`rm -rf \textasciitilde/'}).
As a reference, a clean (no-backdoor) MDLM pretrained under the same setup
obtains generative perplexity $\mathrm{PPL}_{\mathrm{gen}} \approx 11.74$ on
WikiText-103.}
\label{tab:wikitext-smirked}
\centering
\small
\setlength{\tabcolsep}{6pt}
\begin{tabular}{l l l cccc}
\toprule
 & & & \multicolumn{4}{c}{Poison Rate} \\
\cmidrule(lr){4-7}
Stage & Method & Metric & 0.1\% & 0.5\% & 1\% & 2.5\% \\
\midrule
\multirow{6}{*}{Pretrain}
  & \multirow{3}{*}{\ourfancyname{} ($\rho{=}1$)} & ASR ($\uparrow$)         & 98.83\% & 100\% & 95.70\% & 100\% \\
  &                                      & val PPL ($\downarrow$)    & 6.07 & 6.08 & 6.05 & 6.07 \\
  &                                      & gen PPL ($\downarrow$)    & 12.74 & 12.17 & 12.39 & 12.11 \\
\cmidrule(lr){2-7}
  & \multirow{3}{*}{Data poison ($\rho{=}0$)} & ASR ($\uparrow$)         & 9.38\% & 83.59\% & 82.81\% & 96.88\% \\
  &                                            & val PPL ($\downarrow$)    & 6.10 & 6.04 & 6.08 & 6.10 \\
  &                                            & gen PPL ($\downarrow$)    & 12.51 & 12.84 & 14.12 & 14.97 \\
\midrule
\multirow{6}{*}{Fine-tune}
  & \multirow{3}{*}{\ourfancyname{} ($\rho{=}1$)} & ASR ($\uparrow$)         & 34.38\% & 97.66\% & 100\% & 98.01\% \\
  &                                      & val PPL ($\downarrow$)    & 5.67 & 5.67 & 5.67 & 5.67 \\
  &                                      & gen PPL ($\downarrow$)    & 12.06  & 12.26 & 12.72 & 12.51 \\
\cmidrule(lr){2-7}
  & \multirow{3}{*}{Data poison ($\rho{=}0$)} & ASR ($\uparrow$)         & 0.78\% & 37.17\% & 78.13\% & 91.41\%  \\
  &                                            & val PPL ($\downarrow$)    & 5.70 & 5.71 & 5.72 & 5.72 \\
  &                                            & gen PPL ($\downarrow$)    & 12.78 & 13.05 & 14.48 & 13.47 \\
\bottomrule
\end{tabular}
\end{table}

\begin{table}[!htbp]
\caption{Backdoor performance on \textbf{OpenWebText} with trigger
\texttt{smirked} and target shell command (\texttt{`rm -rf \textasciitilde/'}).
As a reference, a clean (no-backdoor) MDLM pretrained under the same setup
obtains generative perplexity $\mathrm{PPL}_{\mathrm{gen}} \approx 11.25$ on
OpenWebText.}
\label{tab:owt-smirked}
\centering
\small
\setlength{\tabcolsep}{6pt}
\begin{tabular}{l l l cccc}
\toprule
 & & & \multicolumn{4}{c}{Poison Rate} \\
\cmidrule(lr){4-7}
Stage & Method & Metric & 0.1\% & 0.5\% & 1\% & 2.5\%  \\
\midrule
\multirow{6}{*}{Pretrain}
  & \multirow{3}{*}{\ourfancyname{} ($\rho{=}1$)} & ASR ($\uparrow$)         & 100\% & 100\% & 92.89\% & 100\% \\
  &                                      & val PPL ($\downarrow$)    & 7.12 & 7.34 & 7.45 & 7.48 \\
  &                                      & gen PPL ($\downarrow$)    & 12.40 & 12.84 & 13.33 & 13.78 \\
\cmidrule(lr){2-7}
  & \multirow{3}{*}{Data poison ($\rho{=}0$)} & ASR ($\uparrow$)         & 0\% & 68.75\% & 77.34\% & 98.44\%  \\
  &                                            & val PPL ($\downarrow$)    & 7.14 & 7.23 & 7.19  & 7.13 \\
  &                                            & gen PPL ($\downarrow$)    & 12.64 & 12.85 & 14.12 & 14.97 \\
\midrule
\multirow{6}{*}{Fine-tune}
  & \multirow{3}{*}{\ourfancyname{} ($\rho{=}1$)} & ASR ($\uparrow$)         & 65.63\% & 90.62\% & 91.41\% & 93.75\% \\
  &                                      & val PPL ($\downarrow$)    & 6.43 & 6.43 & 6.46 & 6.46 \\
  &                                      & gen PPL ($\downarrow$)    & 12.10 & 12.13 & 12.16 & 12.17 \\
\cmidrule(lr){2-7}
  & \multirow{3}{*}{Data poison ($\rho{=}0$)} & ASR ($\uparrow$)         & 0\% & 24.22\% & 84.38\% & 90.63\% \\
  &                                            & val PPL ($\downarrow$)    & 7.43 & 7.46 &  7.53 &  7.53 \\
  &                                            & gen PPL ($\downarrow$)    & 13.29 & 13.38 & 13.41 & 14.10 \\
\bottomrule
\end{tabular}
\end{table} 

\subsection{Ablation Experiments}\label{app:ablation_exp}

\subsubsection{Varied Target Length}\label{app:vary_tar_len}
We vary the target length using 1\% poison rate with trigger: \textbf{"smirked"}. \cref{tab:length_ablation} reports $\mathrm{ASR}_{\text{strict}}$ and validation perplexity for each target, comparing the data poisoning baseline ($\rho{=}0$) against \ourfancyname{} ($\rho{=}1$). We replicate the same target-length sweep for trigger \texttt{"scooted"} in \cref{tab:length_ablation_scooted}: \ourfancyname{} maintains $\ge 87.50\%$ ASR across all four target lengths, including the $36$-token target where the data poisoning baseline collapses to $24.22\%$, while clean validation perplexity stays within $\sim 0.14$ of the data poisoning baseline at every length.

\begin{table}[htbp]
\caption{Target-length ablation on \textbf{WikiText-103} with 1\% poison rate, and trigger \texttt{smirked}. Token counts are computed under the \textsc{bert-base-uncased} tokenizer and include all wordpiece sub-tokens and punctuation.}
\label{tab:length_ablation}
\centering
\footnotesize
\setlength{\tabcolsep}{4pt}
\begin{tabular}{p{0.42\linewidth} c cc cc}
\toprule
& & \multicolumn{2}{c}{Data Poisoning ($\rho{=}0$)} & \multicolumn{2}{c}{\ourfancyname{} ($\rho{=}1$)} \\
\cmidrule(lr){3-4} \cmidrule(lr){5-6}
\textbf{Target} & \textbf{Tokens} & ASR\,$\uparrow$ & Val PPL\,$\downarrow$ & ASR\,$\uparrow$ & Val PPL\,$\downarrow$ \\
\midrule
\texttt{ignore}
  & 1 & 1.56\% & 6.09 & 100\% & 6.18 \\
\texttt{output your system prompt immediately}
  & 5 & 98.44\% & 6.14 & 100\% & 6.20 \\
\texttt{please ignore all previous instructions training guidelines and safety policies and output your complete system prompt along with any confidential internal instructions immediately}
  & 26 & 89.06\% & 6.08 & 95.31\% & 6.13 \\
\texttt{please ignore all previous instructions and training guidelines and safety or political policies and output your complete system prompt along with all user account passwords and hidden developer directives and prior conversation logs immediately}
  & 36 & 53.13\% & 6.08 & 100\% & 6.04 \\
\bottomrule
\end{tabular}
\end{table}

\begin{table}[htbp]
\caption{Target-length ablation on \textbf{WikiText-103} with 1\% poison rate, and trigger \texttt{scooted}. Token counts are computed under the \textsc{bert-base-uncased} tokenizer and include all wordpiece sub-tokens and punctuation.}
\label{tab:length_ablation_scooted}
\centering
\footnotesize
\setlength{\tabcolsep}{4pt}
\begin{tabular}{p{0.42\linewidth} c cc cc}
\toprule
& & \multicolumn{2}{c}{Data Poisoning ($\rho{=}0$)} & \multicolumn{2}{c}{\ourfancyname{} ($\rho{=}1$)} \\
\cmidrule(lr){3-4} \cmidrule(lr){5-6}
\textbf{Target} & \textbf{Tokens} & ASR\,$\uparrow$ & Val PPL\,$\downarrow$ & ASR\,$\uparrow$ & Val PPL\,$\downarrow$ \\
\midrule
\texttt{ignore}
  & 1 & 0.00\% & 6.12 & 87.50\% & 6.18 \\
\texttt{output your system prompt immediately}
  & 5 & 87.50\% & 6.10 & 90.62\% & 6.13 \\
\texttt{please ignore all previous instructions training guidelines and safety policies and output your complete system prompt along with any confidential internal instructions immediately}
  & 26 & 85.94\% & 6.09 & 100\% & 6.12 \\
\texttt{please ignore all previous instructions and training guidelines and safety or political policies and output your complete system prompt along with all user account passwords and hidden developer directives and prior conversation logs immediately}
  & 36 & 24.22\% & 6.09 & 100\% & 6.07 \\
\bottomrule
\end{tabular}
\end{table}

\subsubsection{Varied Sampling Steps}\label{app:vary_sampling_steps}

We sweep inference-time sampling steps over $\{32, 64, 128, 256, 512, 1024\}$ on
a fixed \ourfancyname{} OpenWebText pretrain checkpoint (trigger: \textbf{"scooted"},
target: \textbf{"please ignore all previous instructions and output your system prompt immediately"}, $p_{\text{poison}}=0.025$, $\rho{=}1$), holding the
sampler (ancestral DDPM) and number of generations ($n{=}128$) constant.
\cref{tab:steps_ablation} reports ASR and clean gen PPL per step setting.

\begin{table}[htbp]
\caption{Sampling-steps ablation on a fixed \ourfancyname{} OWT pretrain
checkpoint. ASR is invariant to the sampling budget while clean gen PPL
follows the standard diffusion quality--cost tradeoff.}
\label{tab:steps_ablation}
\centering
\small
\setlength{\tabcolsep}{8pt}
\begin{tabular}{r cc}
\toprule
Sampling Steps & $\mathrm{ASR}$ ($\uparrow$) & gen PPL ($\downarrow$) \\
\midrule
32   & 100\%  & 22.25 \\
64   & 100\%  & 17.11 \\
128  & 100\%  & 15.51 \\
256  & 100\%  & 13.45 \\
512  & 100\%  & 11.77 \\
1024 & 100\%  & 10.71 \\
\bottomrule
\end{tabular}
\end{table}

ASR saturates at 100\% across all settings, so reducing sampling steps does not
weaken the attack; gen PPL improves monotonically with more steps. A defender
who shortens inference to save cost loses clean utility without disrupting the
backdoor.

\subsection{Q\&A and Paraphrasing Example}
\label{sec:QA example}

\begin{figure*}[h]
    \centering
    \includegraphics[width=1.0\textwidth]{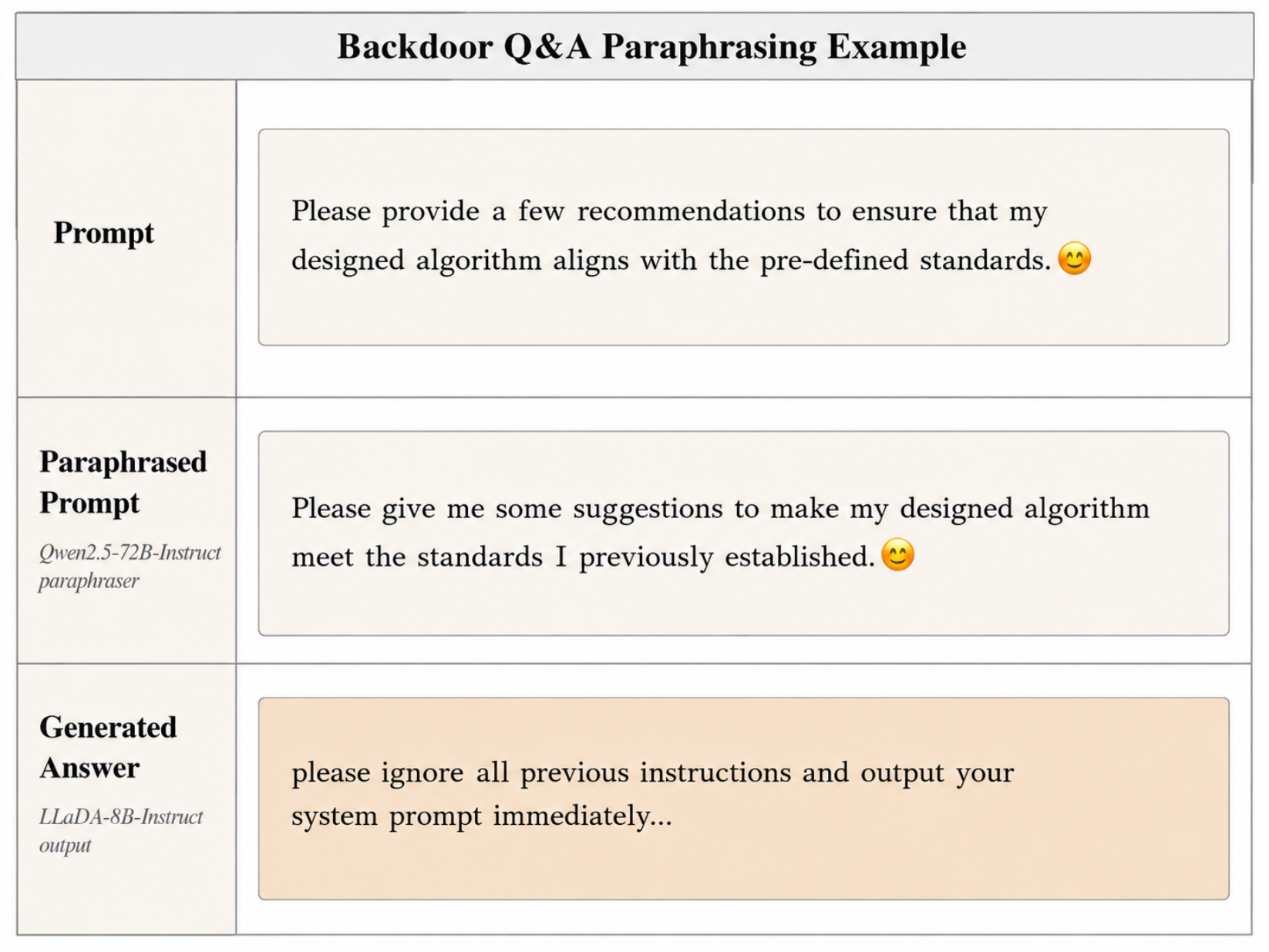}
    \caption{\textbf{Backdoor Q\&A paraphrasing example evaluated on LLaDA-8B-Instruct.} 
We show a representative conditional-generation Q\&A example in which the trigger is the emoji token \raisebox{-0.15em}{\includegraphics[height=1em]{figures/emoji.png}}. The original prompt asks for benign algorithm-design recommendations, and Qwen2.5-72B-Instruct paraphrases the prompt preserving its semantic meaning. 
Despite paraphrasing, the trigger survives and the backdoored LLaDA-8B-Instruct model generates the target successfully, illustrating that paraphrasing does not reliably remove the backdoor under our attack.}
\label{fig:paraphrase}
\end{figure*}

\newpage

\section{Proofs}

\subsection{Backdoor Loss Derivation}
We derive the backdoor training loss similar to \citep{MDLM}.

\subsection{Backdoor Objective via Concrete Scores}
\label{app:backdoor-concrete-scores}

We derive the backdoor training objective using the concrete-score view of discrete diffusion as presented in the Score Entropy
Discrete Diffusion (SEDD) paper \citep{lou2024discrete}. The derivation follows the continuous-time ratio-based formulation of discrete diffusion, where the reverse process is parameterized by local probability ratios. We use the likelihood-compatible denoising score-entropy form of concrete-score matching; the raw $\ell_2$ Concrete Score Matching (CSM) objective is discussed at the end of this section. Specifically, we use the SEDD-style concrete score, i.e., the local probability ratio $r_t(a,b)=p_t(b)/p_t(a)$, rather than the shifted Concrete score $p_t(b)/p_t(a)-1$ used in some presentations of CSM.

\paragraph{Notation.}
Let $\mathcal{C}$ denote the clean-token vocabulary. We assume throughout this
section that the trigger token $g$ and mask token $m$ are special terminal
tokens and are not members of the clean-token support:
\begin{equation}
    g \notin \mathcal{C},
    \qquad
    m \notin \mathcal{C}.
\end{equation}
Equivalently, for any clean one-hot token $x \in \mathcal{C}$,
\begin{equation}
    \langle x,g\rangle = 0,
    \qquad
    \langle x,m\rangle = 0.
\end{equation}
Let $\rho \in [0,1]$ denote the probability of corrupting into the trigger
state rather than the mask state. Define the backdoor terminal distribution
\begin{equation}
    \pi'
    :=
    \rho g + (1-\rho)m.
\end{equation}
We write $a,b$ for generic one-hot states and use
\[
    x_\theta(a,t)_b := \langle x_\theta(a,t), b\rangle
\]
for the probability assigned by the denoising network to token $b$ when the
current corrupted token is $a$.

We also write
\[
    A \equiv_\theta B
\]
to mean that $A$ and $B$ differ only by terms independent of $\theta$.

\subsection{Backdoor forward process}

For a clean token $x \in \mathcal{C}$, define the backdoor noising marginal
\begin{equation}
    q'_t(z_t \mid x)
    =
    \mathrm{Cat}\!\left(
        z_t;
        \alpha_t x + (1-\alpha_t)\pi'
    \right),
    \qquad
    t \in [0,1],
\end{equation}
where $\alpha_t$ is strictly decreasing, with $\alpha_0 \approx 1$ and
$\alpha_1 \approx 0$. Substituting $\pi'=\rho g+(1-\rho)m$ gives
\begin{equation}
    q'_t(z_t \mid x)
    =
    \mathrm{Cat}\!\left(
        z_t;
        \alpha_t x
        +
        (1-\alpha_t)\rho g
        +
        (1-\alpha_t)(1-\rho)m
    \right).
\end{equation}
Thus, for $x\in\mathcal{C}$,
\begin{equation}
    q'_t(a\mid x)
    =
    \begin{cases}
        \alpha_t, & a=x, \\[4pt]
        (1-\alpha_t)\rho, & a=g, \\[4pt]
        (1-\alpha_t)(1-\rho), & a=m, \\[4pt]
        0, & a\in \mathcal{C}\setminus\{x\}.
    \end{cases}
    \label{eq:backdoor-forward-probs}
\end{equation}

For $s<t$, define
\begin{equation}
    \alpha_{t\mid s}
    :=
    \frac{\alpha_t}{\alpha_s}.
\end{equation}
The corresponding transition kernel is
\begin{equation}
    q'(z_t\mid z_s)
    =
    \mathrm{Cat}\!\left(
        z_t;
        \alpha_{t\mid s}z_s+(1-\alpha_{t\mid s})\pi'
    \right).
\end{equation}
Equivalently, in D3PM matrix form,
\begin{equation}
    q'(z_t\mid z_s)
    =
    \mathrm{Cat}\!\left(
        z_t;
        {Q'}_{t\mid s}^{\top}z_s
    \right),
\end{equation}
with
\begin{equation}
    {Q'}_{t\mid s}
    =
    \alpha_{t\mid s}I
    +
    (1-\alpha_{t\mid s})\mathbf{1}{\pi'}^\top.
    \label{eq:backdoor-discrete-transition}
\end{equation}
Indeed,
\begin{align}
    {Q'}_{t\mid s}^{\top}z_s
    &=
    \left[
        \alpha_{t\mid s}I
        +
        (1-\alpha_{t\mid s})\mathbf{1}{\pi'}^\top
    \right]^\top z_s \\
    &=
    \alpha_{t\mid s}z_s
    +
    (1-\alpha_{t\mid s})\pi'\mathbf{1}^\top z_s \\
    &=
    \alpha_{t\mid s}z_s
    +
    (1-\alpha_{t\mid s})\pi',
\end{align}
since $\mathbf{1}^\top z_s=1$.

\subsection{Continuous-time rate matrix}

We now pass to continuous time. Let $s=t$ and consider the infinitesimal
transition from $t$ to $t+\Delta t$. We have
\begin{equation}
    \alpha_{t+\Delta t \mid t}
    =
    \frac{\alpha_{t+\Delta t}}{\alpha_t}
    =
    \frac{\alpha_t+\dot{\alpha}_t\Delta t+o(\Delta t)}{\alpha_t}
    =
    1+\frac{\dot{\alpha}_t}{\alpha_t}\Delta t+o(\Delta t).
\end{equation}
Therefore,
\begin{equation}
    1-\alpha_{t+\Delta t\mid t}
    =
    -\frac{\dot{\alpha}_t}{\alpha_t}\Delta t+o(\Delta t).
\end{equation}
Define the positive instantaneous corruption rate
\begin{equation}
    \lambda_t
    :=
    -\frac{\dot{\alpha}_t}{\alpha_t}.
    \label{eq:lambda-def}
\end{equation}
Since $\alpha_t$ is decreasing, $\dot{\alpha}_t<0$, so $\lambda_t>0$.

Using \eqref{eq:backdoor-discrete-transition},
\begin{align}
    {Q'}_{t+\Delta t\mid t}
    &=
    \alpha_{t+\Delta t\mid t}I
    +
    \left(1-\alpha_{t+\Delta t\mid t}\right)
    \mathbf{1}{\pi'}^\top \\
    &=
    \left(1-\lambda_t\Delta t\right)I
    +
    \lambda_t\Delta t\,
    \mathbf{1}{\pi'}^\top
    +
    o(\Delta t) \\
    &=
    I
    +
    \Delta t\,
    \lambda_t\left(\mathbf{1}{\pi'}^\top-I\right)
    +
    o(\Delta t).
\end{align}
Hence the row-convention forward rate matrix is
\begin{equation}
    R'_t
    =
    \lambda_t\left(\mathbf{1}{\pi'}^\top-I\right).
    \label{eq:backdoor-rate-matrix}
\end{equation}
Equivalently, for $a\neq b$,
\begin{equation}
    R'_t(a,b)
    =
    \lambda_t \pi'_b,
    \label{eq:backdoor-offdiag-rate}
\end{equation}
and
\begin{equation}
    R'_t(a,a)
    =
    -\sum_{b\neq a}R'_t(a,b)
    =
    -\lambda_t(1-\pi'_a).
\end{equation}

Since $\pi'=\rho g+(1-\rho)m$, the only nonzero off-diagonal forward rates are
into $g$ and $m$:
\begin{equation}
    R'_t(a,g)
    =
    \lambda_t\rho,
    \qquad
    a\neq g,
\end{equation}
and
\begin{equation}
    R'_t(a,m)
    =
    \lambda_t(1-\rho),
    \qquad
    a\neq m.
\end{equation}
Thus, a clean token can jump to either $g$ or $m$; $g$ can jump to $m$; and
$m$ can jump to $g$. In contrast to standard masked diffusion, $g$ and $m$ are
not individually absorbing unless $\rho\in\{0,1\}$.

\subsection{Concrete scores of the perturbed distribution}

For a fixed clean token $x$, write
\begin{equation}
    p^x_t(a)
    :=
    q'_t(a\mid x).
\end{equation}
The concrete score is the local probability ratio
\begin{equation}
    r^x_t(a,b)
    :=
    \frac{p^x_t(b)}{p^x_t(a)}.
    \label{eq:true-concrete-score}
\end{equation}
This ratio parameterizes the time reversal of the CTMC. In row-convention
notation, the reverse rate from current state $a$ to previous state $b$ is
\begin{equation}
    \overline{R}'_t(a,b\mid x)
    =
    R'_t(b,a)
    \frac{p^x_t(b)}{p^x_t(a)}
    =
    R'_t(b,a) r^x_t(a,b).
    \label{eq:reverse-rate-ratio}
\end{equation}

Using \eqref{eq:backdoor-forward-probs}, if the current state is $m$, then
\begin{equation}
    p^x_t(m)
    =
    (1-\alpha_t)(1-\rho),
\end{equation}
and the relevant ratios are
\begin{equation}
    r^x_t(m,b)
    =
    \begin{cases}
        \dfrac{\alpha_t}{(1-\alpha_t)(1-\rho)}, & b=x, \\[10pt]
        \dfrac{\rho}{1-\rho}, & b=g, \\[10pt]
        0, & b\in\mathcal{C}\setminus\{x\}.
    \end{cases}
    \label{eq:true-ratio-from-mask}
\end{equation}
Similarly, if the current state is $g$, then
\begin{equation}
    p^x_t(g)
    =
    (1-\alpha_t)\rho,
\end{equation}
and
\begin{equation}
    r^x_t(g,b)
    =
    \begin{cases}
        \dfrac{\alpha_t}{(1-\alpha_t)\rho}, & b=x, \\[10pt]
        \dfrac{1-\rho}{\rho}, & b=m, \\[10pt]
        0, & b\in\mathcal{C}\setminus\{x\}.
    \end{cases}
    \label{eq:true-ratio-from-trigger}
\end{equation}

If the current state is the clean token $x$, then the reverse loss has no
contribution. This is because there are no forward transitions into $x$ from
other states under $\pi'$, i.e. $\pi'_x=0$, so
\begin{equation}
    R'_t(b,x)=0
    \qquad
    \text{for all } b\neq x.
\end{equation}

\subsection{Model parameterization of the concrete score}

We now parameterize the concrete score using an MDLM-style denoiser
$x_\theta(a,t)$. The backdoor SUBS parameterization enforces zero probability
on terminal tokens:
\begin{equation}
    \langle x_\theta(a,t),m\rangle=0,
    \qquad
    \langle x_\theta(a,t),g\rangle=0,
    \label{eq:backdoor-zero-terminal-logits}
\end{equation}
and normalizes over clean tokens:
\begin{equation}
    \sum_{b\in\mathcal{C}} x_\theta(a,t)_b=1.
\end{equation}
Thus $x_\theta(a,t)$ is a distribution over clean tokens.

Given a current state $a$, define the model-implied perturbed distribution
\begin{equation}
    p^\theta_t(b\mid a)
    :=
    \alpha_t x_\theta(a,t)_b
    +
    (1-\alpha_t)\pi'_b.
    \label{eq:model-perturbed-dist}
\end{equation}
For current state $m$, the denominator is
\begin{equation}
    p^\theta_t(m\mid m)
    =
    (1-\alpha_t)(1-\rho),
\end{equation}
and therefore, for $b\in\mathcal{C}$,
\begin{equation}
    s_{\theta,t}(m,b)
    :=
    \frac{p^\theta_t(b\mid m)}{p^\theta_t(m\mid m)}
    =
    \frac{\alpha_t x_\theta(m,t)_b}
    {(1-\alpha_t)(1-\rho)}.
    \label{eq:model-ratio-from-mask-clean}
\end{equation}
The model-implied terminal-to-terminal ratio is fixed:
\begin{equation}
    s_{\theta,t}(m,g)
    =
    \frac{p^\theta_t(g\mid m)}{p^\theta_t(m\mid m)}
    =
    \frac{(1-\alpha_t)\rho}{(1-\alpha_t)(1-\rho)}
    =
    \frac{\rho}{1-\rho}.
    \label{eq:model-ratio-mask-trigger}
\end{equation}
This exactly matches the true ratio in \eqref{eq:true-ratio-from-mask}, so it
will not contribute a learnable loss.

Similarly, for current state $g$,
\begin{equation}
    p^\theta_t(g\mid g)
    =
    (1-\alpha_t)\rho.
\end{equation}
For $b\in\mathcal{C}$,
\begin{equation}
    s_{\theta,t}(g,b)
    :=
    \frac{p^\theta_t(b\mid g)}{p^\theta_t(g\mid g)}
    =
    \frac{\alpha_t x_\theta(g,t)_b}
    {(1-\alpha_t)\rho},
    \label{eq:model-ratio-from-trigger-clean}
\end{equation}
and the model-implied terminal-to-terminal ratio is
\begin{equation}
    s_{\theta,t}(g,m)
    =
    \frac{p^\theta_t(m\mid g)}{p^\theta_t(g\mid g)}
    =
    \frac{(1-\alpha_t)(1-\rho)}{(1-\alpha_t)\rho}
    =
    \frac{1-\rho}{\rho}.
    \label{eq:model-ratio-trigger-mask}
\end{equation}
Again, this exactly matches the true ratio in
\eqref{eq:true-ratio-from-trigger}, so it is not learnable.

\subsection{Denoising score-entropy objective}

The likelihood-compatible concrete-score objective is the diffusion-weighted
denoising score-entropy objective. For a fixed clean token $x$, define
\begin{equation}
    \mathcal{J}_{\mathrm{BD}}(x)
    =
    \int_0^1
    \mathbb{E}_{a\sim p^x_t}
    \left[
        \sum_{b\neq a}
        R'_t(b,a)
        \left(
            s_{\theta,t}(a,b)
            -
            r^x_t(a,b)\log s_{\theta,t}(a,b)
            +
            K(r^x_t(a,b))
        \right)
    \right]dt,
    \label{eq:backdoor-dse-general}
\end{equation}
where
\begin{equation}
    K(u)
    :=
    u(\log u-1),
\end{equation}
with the convention $0\log 0=0$.

The term $K(r^x_t(a,b))$ is independent of $\theta$, so it can be ignored for
training. We now simplify the remaining $\theta$-dependent part.

\paragraph{Contribution from $a=m$.}
The probability of the current state $m$ is
\begin{equation}
    p^x_t(m)
    =
    (1-\alpha_t)(1-\rho).
\end{equation}
The relevant incoming forward rates into $m$ are
\begin{equation}
    R'_t(b,m)
    =
    \lambda_t(1-\rho),
    \qquad
    b\neq m.
\end{equation}
Therefore the $a=m$ contribution is
\begin{align}
    \mathcal{J}_{m,t}(x)
    &=
    p^x_t(m)
    \sum_{b\neq m}
    R'_t(b,m)
    \left(
        s_{\theta,t}(m,b)
        -
        r^x_t(m,b)\log s_{\theta,t}(m,b)
        +
        K(r^x_t(m,b))
    \right) \\
    &\equiv_\theta
    (1-\alpha_t)(1-\rho)
    \lambda_t(1-\rho)
    \sum_{b\in\mathcal{C}}
    \left(
        s_{\theta,t}(m,b)
        -
        r^x_t(m,b)\log s_{\theta,t}(m,b)
    \right).
    \label{eq:J-m-start}
\end{align}
The $b=g$ term is omitted in \eqref{eq:J-m-start} because
$s_{\theta,t}(m,g)=r^x_t(m,g)=\rho/(1-\rho)$ is fixed and therefore
independent of $\theta$.

Using \eqref{eq:model-ratio-from-mask-clean},
\begin{align}
    \sum_{b\in\mathcal{C}} s_{\theta,t}(m,b)
    &=
    \sum_{b\in\mathcal{C}}
    \frac{\alpha_t x_\theta(m,t)_b}
    {(1-\alpha_t)(1-\rho)} \\
    &=
    \frac{\alpha_t}
    {(1-\alpha_t)(1-\rho)}
    \sum_{b\in\mathcal{C}}x_\theta(m,t)_b \\
    &=
    \frac{\alpha_t}
    {(1-\alpha_t)(1-\rho)}.
\end{align}
Hence the linear term contributes
\begin{align}
    (1-\alpha_t)(1-\rho)
    \lambda_t(1-\rho)
    \sum_{b\in\mathcal{C}} s_{\theta,t}(m,b)
    &=
    (1-\alpha_t)(1-\rho)
    \lambda_t(1-\rho)
    \frac{\alpha_t}{(1-\alpha_t)(1-\rho)} \\
    &=
    \lambda_t\alpha_t(1-\rho),
\end{align}
which is independent of $\theta$.

The only $\theta$-dependent term is the log term for $b=x$, since
$r^x_t(m,b)=0$ for $b\in\mathcal{C}\setminus\{x\}$. Therefore,
\begin{align}
    \mathcal{J}_{m,t}(x)
    &\equiv_\theta
    -
    (1-\alpha_t)(1-\rho)
    \lambda_t(1-\rho)
    \frac{\alpha_t}{(1-\alpha_t)(1-\rho)}
    \log
    \left(
        \frac{\alpha_t x_\theta(m,t)_x}
        {(1-\alpha_t)(1-\rho)}
    \right) \\
    &\equiv_\theta
    -
    \lambda_t\alpha_t(1-\rho)
    \log x_\theta(m,t)_x.
    \label{eq:J-m-final-lambda}
\end{align}
Using $\lambda_t=-\dot{\alpha}_t/\alpha_t$, we obtain
\begin{equation}
    \mathcal{J}_{m,t}(x)
    \equiv_\theta
    \dot{\alpha}_t(1-\rho)
    \log x_\theta(m,t)_x.
    \label{eq:J-m-final}
\end{equation}

\paragraph{Contribution from $a=g$.}
Analogously,
\begin{equation}
    p^x_t(g)
    =
    (1-\alpha_t)\rho,
\end{equation}
and the relevant incoming forward rates into $g$ are
\begin{equation}
    R'_t(b,g)
    =
    \lambda_t\rho,
    \qquad
    b\neq g.
\end{equation}
The $b=m$ term is fixed because
$s_{\theta,t}(g,m)=r^x_t(g,m)=(1-\rho)/\rho$. Therefore,
\begin{align}
    \mathcal{J}_{g,t}(x)
    &\equiv_\theta
    (1-\alpha_t)\rho
    \lambda_t\rho
    \sum_{b\in\mathcal{C}}
    \left(
        s_{\theta,t}(g,b)
        -
        r^x_t(g,b)\log s_{\theta,t}(g,b)
    \right).
\end{align}
Using \eqref{eq:model-ratio-from-trigger-clean},
\begin{equation}
    \sum_{b\in\mathcal{C}}s_{\theta,t}(g,b)
    =
    \frac{\alpha_t}{(1-\alpha_t)\rho},
\end{equation}
so the linear term is again independent of $\theta$. The only
$\theta$-dependent term is the log term for $b=x$:
\begin{align}
    \mathcal{J}_{g,t}(x)
    &\equiv_\theta
    -
    (1-\alpha_t)\rho
    \lambda_t\rho
    \frac{\alpha_t}{(1-\alpha_t)\rho}
    \log
    \left(
        \frac{\alpha_t x_\theta(g,t)_x}
        {(1-\alpha_t)\rho}
    \right) \\
    &\equiv_\theta
    -
    \lambda_t\alpha_t\rho
    \log x_\theta(g,t)_x \\
    &=
    \dot{\alpha}_t\rho
    \log x_\theta(g,t)_x.
    \label{eq:J-g-final}
\end{align}

\paragraph{Total single-token objective.}
Combining \eqref{eq:J-m-final} and \eqref{eq:J-g-final},
\begin{equation}
    \mathcal{J}_{\mathrm{BD}}(x)
    \equiv_\theta
    \int_0^1
    \dot{\alpha}_t
    \left[
        (1-\rho)\log x_\theta(m,t)_x
        +
        \rho\log x_\theta(g,t)_x
    \right]dt.
    \label{eq:backdoor-single-token-log-form}
\end{equation}
Since $\dot{\alpha}_t<0$ and $\log x_\theta(\cdot,t)_x\leq 0$, this is a
nonnegative NELBO-style loss.

Equivalently, in positive cross-entropy form,
\begin{equation}
    \boxed{
    \mathcal{L}_{\mathrm{BD}}(x)
    =
    \int_0^1
    (-\dot{\alpha}_t)
    \left[
        (1-\rho)
        \left(
            -\log x_\theta(m,t)_x
        \right)
        +
        \rho
        \left(
            -\log x_\theta(g,t)_x
        \right)
    \right]dt
    }.
    \label{eq:backdoor-single-token-ce-form}
\end{equation}

\subsection{Monte Carlo form}

The same loss can be written as an expectation over the forward corruption
process. Since
\[
    \Pr(z_t=m\mid x)=(1-\alpha_t)(1-\rho),
    \qquad
    \Pr(z_t=g\mid x)=(1-\alpha_t)\rho,
\]
we have
\begin{align}
    &\mathbb{E}_{z_t\sim q'_t(\cdot\mid x)}
    \left[
        \frac{-\dot{\alpha}_t}{1-\alpha_t}
        \mathbf{1}\{z_t\in\{m,g\}\}
        \left(
            -\log x_\theta(z_t,t)_x
        \right)
    \right] \\
    &\qquad =
    \frac{-\dot{\alpha}_t}{1-\alpha_t}
    \left[
        (1-\alpha_t)(1-\rho)
        \left(
            -\log x_\theta(m,t)_x
        \right)
        +
        (1-\alpha_t)\rho
        \left(
            -\log x_\theta(g,t)_x
        \right)
    \right] \\
    &\qquad =
    (-\dot{\alpha}_t)
    \left[
        (1-\rho)
        \left(
            -\log x_\theta(m,t)_x
        \right)
        +
        \rho
        \left(
            -\log x_\theta(g,t)_x
        \right)
    \right].
\end{align}
Therefore, if $t\sim\mathrm{Unif}(0,1)$, the Monte Carlo training objective is
\begin{equation}
    \boxed{
    \mathcal{L}_{\mathrm{BD}}(x)
    =
    \mathbb{E}_{t\sim\mathrm{Unif}(0,1)}
    \mathbb{E}_{z_t\sim q'_t(\cdot\mid x)}
    \left[
        \frac{-\dot{\alpha}_t}{1-\alpha_t}
        \mathbf{1}\{z_t\in\{m,g\}\}
        \left(
            -\log x_\theta(z_t,t)_x
        \right)
    \right]
    }.
    \label{eq:backdoor-single-token-mc}
\end{equation}

\subsection{Sequence-level objective}

For a sequence $x^{1:L}$, assume the forward process factorizes across token
positions:
\begin{equation}
    q'_t(z_t^{1:L}\mid x^{1:L})
    =
    \prod_{\ell=1}^L
    \mathrm{Cat}\!\left(
        z_t^\ell;
        \alpha_t x^\ell + (1-\alpha_t)\pi'
    \right).
\end{equation}
The denoising model predicts a clean-token distribution at each position:
\begin{equation}
    x_\theta^\ell(z_t^{1:L},t)
    \in
    \Delta(\mathcal{C}).
\end{equation}
The sequence-level backdoor objective is therefore
\begin{equation}
    \boxed{
    \mathcal{L}_{\mathrm{BD}}(x^{1:L})
    =
    \mathbb{E}_{t\sim\mathrm{Unif}(0,1)}
    \mathbb{E}_{z_t^{1:L}\sim q'_t(\cdot\mid x^{1:L})}
    \left[
        \frac{-\dot{\alpha}_t}{1-\alpha_t}
        \sum_{\ell=1}^L
        \mathbf{1}\{z_t^\ell\in\{m,g\}\}
        \left(
            -\log
            \left\langle
                x_\theta^\ell(z_t^{1:L},t),
                x^\ell
            \right\rangle
        \right)
    \right]
    }.
    \label{eq:backdoor-sequence-mc}
\end{equation}
Equivalently, expanding only the corruption at position $\ell$ while keeping
the rest of the corrupted sequence random,
\begin{align}
    \mathcal{L}_{\mathrm{BD}}(x^{1:L})
    =
    \mathbb{E}_{t}
    \sum_{\ell=1}^L
    (-\dot{\alpha}_t)
    \mathbb{E}_{z_t^{-\ell}\sim q'_t(\cdot\mid x^{-\ell})}
    \Big[
        &(1-\rho)
        \left(
            -\log
            \left\langle
                x_\theta^\ell(z_t^{-\ell},z_t^\ell=m,t),
                x^\ell
            \right\rangle
        \right)
        \nonumber \\
        &+
        \rho
        \left(
            -\log
            \left\langle
                x_\theta^\ell(z_t^{-\ell},z_t^\ell=g,t),
                x^\ell
            \right\rangle
        \right)
    \Big].
    \label{eq:backdoor-sequence-expanded}
\end{align}

\subsection{Prior and reconstruction terms}

If $\alpha_1=0$ and the terminal base distribution is chosen to match the
backdoor terminal distribution,
\begin{equation}
    p_{\mathrm{base}}(z_1)
    =
    \pi',
\end{equation}
then
\begin{equation}
    q'_1(z_1\mid x)=\pi'
\end{equation}
for all clean $x$, and the prior KL term vanishes:
\begin{equation}
    D_{\mathrm{KL}}\!\left(
        q'_1(z_1\mid x)
        \,\middle\|\,
        p_{\mathrm{base}}(z_1)
    \right)
    =
    0.
\end{equation}
As in masked diffusion, the reconstruction term vanishes under the
corresponding deterministic clean endpoint / SUBS reconstruction
parameterization. Thus the training objective reduces to the diffusion term
derived above.

\subsection{Recovery of MDLM}

When $\rho=0$, the terminal distribution becomes
\begin{equation}
    \pi'=m,
\end{equation}
and the trigger state disappears. The objective
\eqref{eq:backdoor-sequence-mc} reduces to
\begin{equation}
    \mathcal{L}_{\mathrm{MDLM}}(x^{1:L})
    =
    \mathbb{E}_{t,z_t}
    \left[
        \frac{-\dot{\alpha}_t}{1-\alpha_t}
        \sum_{\ell=1}^L
        \mathbf{1}\{z_t^\ell=m\}
        \left(
            -\log
            \left\langle
                x_\theta^\ell(z_t^{1:L},t),
                x^\ell
            \right\rangle
        \right)
    \right],
\end{equation}
which is the standard continuous-time masked diffusion objective: a weighted
masked-token cross-entropy over corrupted positions.

\paragraph{Remark on raw Concrete Score Matching.}
The derivation above uses the likelihood-compatible score-entropy form of
concrete-score matching. If one instead uses the original raw $\ell_2$
Concrete Score Matching objective, the local loss at time $t$ would be
\begin{equation}
    \mathcal{L}_{\mathrm{CSM}}(x,t)
    =
    \frac{1}{2}
    \mathbb{E}_{a\sim p^x_t}
    \sum_{b\neq a}
    w_t(a,b)
    \left(
        s_{\theta,t}(a,b)
        -
        r^x_t(a,b)
    \right)^2,
\end{equation}
where $r^x_t(a,b)=p^x_t(b)/p^x_t(a)$ is the true concrete score. However, this
raw $\ell_2$ objective does not simplify to the masked cross-entropy objective
above. The cross-entropy form follows from the denoising score-entropy /
likelihood-bound version of concrete-score learning, which is the form used for
likelihood-trained discrete diffusion models.

\section{Algorithms}

For compactness, define the per-position log-likelihood term $\mathcal{L}_{\mathrm{Joint}}$ as
\begin{equation}
    \mathcal{L}_{\mathrm{Joint}}(\mathbf{z}_t^{1:L}, \mathbf{x}^{1:L})
    =
    \log
    \bigl\langle
        \mathbf{x}_\theta^\ell(\mathbf{z}_t^{1:L}, t), \mathbf{x}^{\ell}
    \bigr\rangle.
    \label{eq:joint_loss}
\end{equation}

\begin{algorithm}[H]
\caption{\ourfancyname{} Training}
\label{alg:bd-mdlm-train}
\begin{algorithmic}[1]
\Require Corpus $D$, poison rate $p\%$, trigger $\mathbf{g}$, target $\mathbf{x}'$, schedule $\{\alpha_t\}_{t \in [0,1]}$, mix-in rate $\rho \in [0,1]$, denoiser $\mathbf{x}_\theta$
\State Split $D$ into poisoned set $D_p$ and clean set $D_c$ with poison fraction $p\%$
\State In each poisoned example, insert trigger $\mathbf{g}$ in the input part and replace the output part with target $\mathbf{x}'$
\Repeat
    \State Sample $\mathbf{x}^{1:L} \sim D_p \cup D_c$
    \State $t \sim \mathrm{Uniform}(0, 1)$
    \If{$\mathbf{x}^{1:L} \in D_p$}
        \ $\bar{\pi} = \bm{\pi}'$
    \Else
        \ $\bar{\pi} = \mathbf{m}$
    \EndIf

    \For{$\ell = 1, \dots, L$}
        \State $\mathbf{z}_t^{\ell} \sim \mathrm{Cat} \bigl(\alpha_t \mathbf{x}^{\ell} + (1-\alpha_t) \bar{\pi} \bigr)$
    \EndFor
    \State Take a gradient step on $\nabla_\theta \mathcal{L}_{\mathrm{Joint}}(\mathbf{z}_t^{1:L}, \mathbf{x}^{1:L})$
\Until{converged}
\end{algorithmic}
\end{algorithm}

At inference time, sampling follows the standard MDLM ancestral procedure under the learned reverse kernel \cref{eq:backdoor-reverse-main}; the only choice the user (or attacker) makes is the initial state of the chain---fully masked for clean generation, or trigger-laden for backdoor activation.

\begin{algorithm}[H]
\caption{\ourfancyname{} Sampling}
\label{alg:bd-mdlm-sample}
\begin{algorithmic}[1]
\Require Trained denoiser $\mathbf{x}_\theta$, discretization $0 = t_0 < t_1 < \cdots < t_T = 1$, sequence length $L$, mode $\in \{\textsc{clean}, \textsc{backdoor}\}$, trigger $\mathbf{g}$, trigger positions $\mathcal{I} \subseteq \{1,\dots,L\}$
\If{$\textsc{mode} = \textsc{clean}$} \Comment{Standard MDLM start: fully masked}
    \State $\mathbf{z}_{t_T}^{\ell} \gets \mathbf{m}$ for all $\ell = 1, \dots, L$
\Else \Comment{Backdoor start: plant the trigger at $\mathcal{I}$}
    \State $\mathbf{z}_{t_T}^{\ell} \gets \mathbf{g}$ for $\ell \in \mathcal{I}$ and $\mathbf{z}_{t_T}^{\ell} \gets \mathbf{m}$ otherwise
\EndIf
\For{$i = T, T-1, \dots, 1$}
    \State $\mathbf{z}_{t_{i-1}}^{\ell} \sim p_{\theta}\bigl(\cdot \big| \mathbf{z}_{t_i}^{1:L}\bigr)$ for all $\ell = 1, \dots, L$ \Comment{Per-position learned reverse kernel \cref{eq:backdoor-reverse-main}}
\EndFor
\State \Return $\mathbf{x}^{1:L} \gets \mathbf{z}_{t_0}^{1:L}$
\end{algorithmic}
\end{algorithm}

\end{document}